\documentclass[sigconf]{acmart}
\pdfoutput=1

\makeatletter
\def\@ACM@checkaffil{% Only warnings
    \if@ACM@instpresent\else
    \ClassWarningNoLine{\@classname}{No institution present for an affiliation}%
    \fi
    \if@ACM@citypresent\else
    \ClassWarningNoLine{\@classname}{No city present for an affiliation}%
    \fi
    \if@ACM@countrypresent\else
        \ClassWarningNoLine{\@classname}{No country present for an affiliation}%
    \fi
}
\makeatother

\AtBeginDocument{%
  \providecommand\BibTeX{{%
    \normalfont B\kern-0.5em{\scshape i\kern-0.25em b}\kern-0.8em\TeX}}}

\copyrightyear{2024}
\acmYear{2024}
\setcopyright{acmlicensed}
\acmConference[KDD '24] {Proceedings of the 30th ACM SIGKDD Conference on Knowledge Discovery and Data Mining }{August 25--29, 2024}{Barcelona, Spain.}
\acmBooktitle{Proceedings of the 30th ACM SIGKDD Conference on Knowledge Discovery and Data Mining (KDD '24), August 25--29, 2024, Barcelona, Spain}
\acmISBN{979-8-4007-0490-1/24/08}
\acmDOI{10.1145/3637528.3671770}

\usepackage{multirow}
\usepackage{bm}
\usepackage{textcomp}
\usepackage{subfigure}
\usepackage[normalem]{ulem}
\useunder{\uline}{\ul}{}
\usepackage{stfloats}
\usepackage{pifont}
\begin{document}
% \fancyhead{}
\title{Co-Neighbor Encoding Schema: A Light-cost Structure Encoding Method for Dynamic Link Prediction}

% \author{Anonymous Author(s), Submission ID: 100}

\author{Ke Cheng}
\email{ckpassenger@buaa.edu.cn}
\affiliation{%
  \institution{CCSE Lab, Beihang University}
  \city{Beijing}
  \country{China}
}

\author{Linzhi Peng}
\email{lzpeng626@buaa.edu.cn}
\affiliation{%
  \institution{CCSE Lab, Beihang University}
  \city{Beijing}
  \country{China}
}

\author{Junchen Ye}
\email{junchenye@buaa.edu.cn}
\authornote{Corresponding Author.}
\affiliation{%
  \institution{School of Transportation Science and Engineering, Beihang University}
  \city{Beijing}
  \country{China}
}

\author{Leilei Sun}
\email{leileisun@buaa.edu.cn}
\affiliation{%
  \institution{CCSE Lab, Beihang University}
  \city{Beijing}
  \country{China}
}

\author{Bowen Du}
\email{dubowen@buaa.edu.cn}
\affiliation{
  \institution{Zhongguancun Laboratory}
}
\affiliation{
  \institution{School of Transportation Science and Engineering, Beihang University}
  \city{Beijing}
  \country{China}
}

\renewcommand{\shortauthors}{Ke Cheng et al.}

\begin{abstract}

Structure encoding has proven to be the key feature to distinguishing links in a graph. However, Structure encoding in the temporal graph keeps changing as the graph evolves, repeatedly computing such features can be time-consuming due to the high-order subgraph construction. 
We develop the Co-Neighbor Encoding Schema (CNES) to address this issue. 
Instead of recomputing the feature by the link, CNES stores information in the memory to avoid redundant calculations. Besides, unlike the existing memory-based dynamic graph learning method that stores node hidden states, we introduce a hashtable-based memory to compress the adjacency matrix for efficient structure feature construction and updating with vector computation in parallel. 
Furthermore, CNES introduces a Temporal-Diverse Memory to generate long-term and short-term structure encoding for neighbors with different structural information. 
A dynamic graph learning framework, Co-Neighbor Encoding Network (CNE-N), is proposed using the aforementioned techniques. Extensive experiments on thirteen public datasets verify the effectiveness and efficiency of the proposed method.

\end{abstract}

%After computing the structure encoding for each low-order subgraph, we aggregate them to obtain the high-order structure encoding, which is more efficient and suitable for parallel computation. 

\begin{CCSXML}
<ccs2012>
<concept>
<concept_id>10002951.10003227.10003351</concept_id>
<concept_desc>Information systems~Data mining</concept_desc>
<concept_significance>500</concept_significance>
</concept>
</ccs2012>
\end{CCSXML}

\ccsdesc[500]{Information systems~Data mining}


\keywords{Graph Neural Networks, Dynamic Graph, Network Embedding}

\maketitle

\section{Introduction}
\label{section-1}
Temporal graphs denote entities as nodes and represent their interactions as edges with timestamps. This is a powerful way to model the dynamics and evolution of complex systems over time. Researchers have leveraged this approach and built many practical systems in a variety of real-world scenarios such as recommendations in social networks \cite{alvarez2021evolutionary, song2019session, kumar2019predicting},
financial networks \cite{ranshous2015anomaly, wang2021bipartite, chang2021f},user-item interaction systems \cite{li2021dynamic, fan2021continuous, yu2022element, yu2022modelling, yu2021deep}. Dynamic graphs can be represented in two ways: Discrete-time and Continuous-time. Discrete-time Dynamic Graphs (DTDG) are sequences of static graph snapshots captured at regular intervals. In contrast, Continuous-time Dynamic Graphs (CTDG) are represented as timed lists of events that include edge or node addition or deletion. CTDGs offer better flexibility and performance than DTDGs and speed up computation by estimating the full graph encoding with a neighborhood subgraph.
%Correspondingly, there are two main strategies in the CTDG method to construct the local temporal subgraph; the first strategy is to extract the subgraph by neighbor sampling; the method can construct a precise subgraph, but neighbor sampling can not be computed in GPU in parallel, which result in low effectiveness, especially when extracting subgraph with 2 or more hops. Another strategy stores the node's history interaction in a memory layer and updates the memory by the graph evolves\cite{kumar2019predicting}, the method can generate node representation without explicit subgraph construction but suffers from a high error rate because the memory stores the history snapshot of the temporal graph rather than the up-to-date temporal graph.

%Temporal graphs denote entities as nodes and represent their interactions as edges with timestamps. This is a powerful way to model the dynamics and evolution of complex systems over time. Dynamic graphs can be represented in two ways: Discrete-time and Continuous-time. Discrete-time Dynamic Graphs (DTDG) are sequences of static graph snapshots captured at regular intervals. In contrast, Continuous-time Dynamic Graphs (CTDG) are represented as timed lists of events that include edge or node addition or deletion, and node or edge feature transformations. CTDGs offer better flexibility and performance than DTDGs. 

% previous works can not encode structure feature

\begin{figure}[t]
\centering
\includegraphics[width=0.9\linewidth]{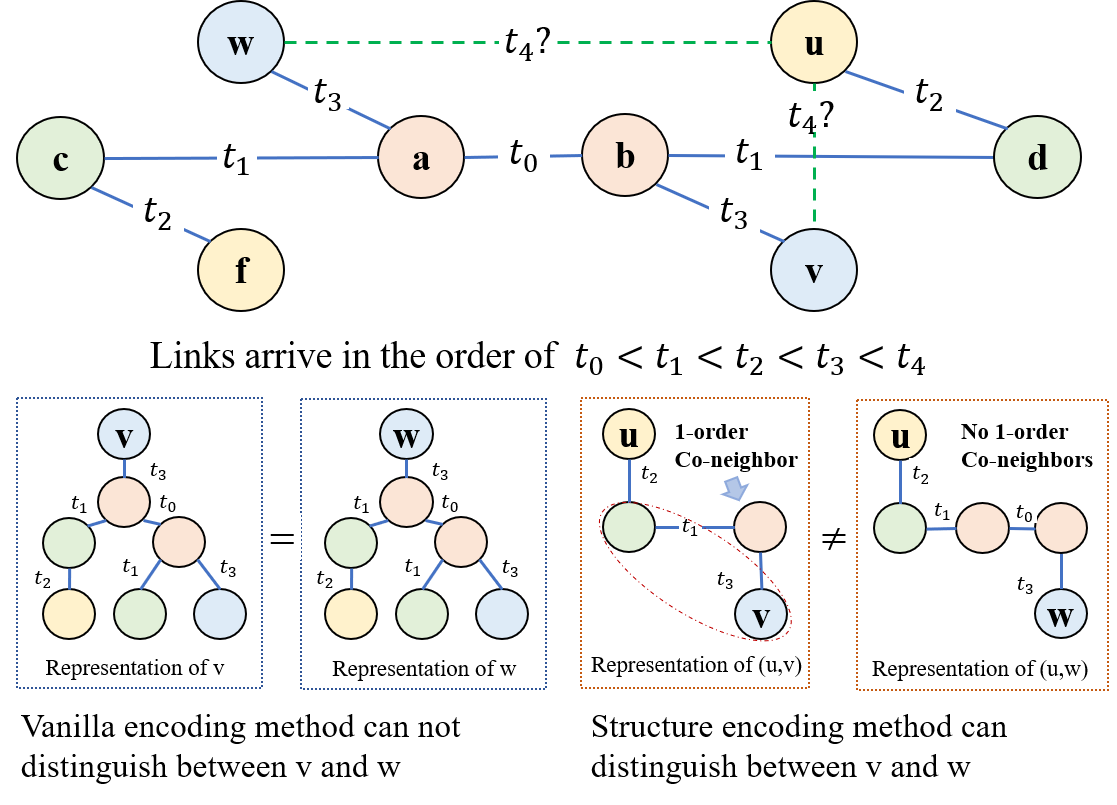}
\caption{An example to show the difference between methods w/o structure encoding, vanilla temporal graph learning methods can not distinguish between link (u,v) and (v,w) while structure encoding-based methods can show the difference by counting neighborhood overlap co-neighbors.}
\vspace{-0.5cm}
\label{Fig:example}
\end{figure}

According to whether to use heuristic features, existing CTDG works can be divided into two main methods to encode the dynamic graph: the vanilla method and the structure encoding-based method.  We discuss the difference between the two methods in Figure \ref{Fig:example}. The vanilla method individually encodes the node neighborhood and concentrates more on the encoding of temporal information. For example, Jodie\cite{kumar2019predicting} models how a user's preferences evolve with an RNN-based encoder that updates the node representation as the user-item interaction occurs. GraphMixer\cite{cong2023we}, and TCL\cite{wang2021tcl} discuss the benefits of capturing the temporal correlation between nodes in the historical interaction sequence with an MLPMixer encoder or a Transformer encoder. However, these methods with only one anchor node treat neighborhood structure as a way to aggregate temporal information, ignoring the structural information itself, which results in the loss of rich information in high-order structures such as relative position information between nodes.

In contrast, the structure encoding-based method encodes the link by generating query-specific encoding of the two end nodes\cite{linkPredictionBasedOnGNN, labelingTrickTheoryOfUsingGNNForMultiNodeRepLearning}, which is to encode one node's neighborhood condition on another node, the method can increase model express ability by better distinguish node neighborhood encoding with the heuristic structure feature like the number of common neighbors.
For example, CAWN\cite{xu2020inductive} extracts several random walk sequences starting from each end node and defines the relative position label as the node position in the sequences. DyGFormer\cite{yu2023towards} extracts the first hop neighbor sequence of the end nodes and computes the neighbor co-occurrence encoding as the first-order relative position label. 
The effectiveness of the structure encoding is largely influenced by the size of the neighborhood to be encoded, a larger neighborhood subgraph will lead to a higher encoding accuracy as well as time complexity\cite{PINT2022}.

\begin{table*}[t]
\caption{ Summary of different CTDG models. We also list the time complexity (per sample, the first element before + stands for sample complexity) and space complexity (per graph, $-$ stands for no memory to store) for each algorithm, with hidden dimension $d$, neighbor sample length $L$, neighbor sample hops $k$, nodes number $N$, node memory/feature dim $M$, we assume all the model shares a single layer full connected neural network as feature encoding model, detailed discussion can be found in Section \ref{sec:complexity}.}
\label{tab:compare method}
\resizebox{11cm}{!}{
\centering
\begin{tabular}{l|cc|cccc|c}
\hline
Model             & Jodie\cite{kumar2019predicting}     & TGN\cite{DBLP:journals/corr/abs-2006-10637}       & CAWN\cite{xu2020inductive}     & DyGFormer\cite{yu2023towards} & NAT\cite{luo2022neighborhood}       & PINT\cite{PINT2022}      & CNE-N    \\ \hline
Strucute Encoding & \ding{55} & \ding{55} & \ding{52} & \ding{52} & \ding{52} & \ding{52} & \ding{52} \\
Lightcost Construction    & \ding{52} & \ding{55} & \ding{55} & \ding{52} & \ding{52} & \ding{55} & \ding{52} \\
Parallel Computation    & \ding{52} & \ding{52} & \ding{55} & \ding{55} & \ding{52} & \ding{52} & \ding{52} \\
Temporal Diversity    & \ding{55} & \ding{55} & \ding{55} & \ding{55} & \ding{55} & \ding{55} & \ding{52} \\ \hline
Time              &     $Md$     &     $L^{k} + L^{k}Md$     &     $L^{k} +  L^{k}Md $    &   $L +   L(L+Md)$      &     $LMd$    &    $L^{k} +  L^{k}Nd$     &   $L + LMd$       \\
Space             &    $NM$     &      $NM$     &     -      &       -    &     $NM$      &     $N^2$      &      $NM$     \\ \hline
\end{tabular}
}
\vspace{-0.3cm}
\end{table*}

%(2) \textit{Memory-based model without neighbor sampling faces high error rate}: Memory-based models without neighbor sampling can not extract up-to-date subgraph. As a result, the model can not precisely gather historical information from neighbors and deliver information about the new-occurring link to them. Besides, the method can hardly be applied to an inductive setting as the memory can not be built up for the unseen nodes; 

Previous methods for generating structure encoding have three drawbacks. 
Firstly, \textbf{constructing a neighborhood subgraph is inefficient}. 
Existing structure encoding is generated by constructing a high-order subgraph assigning relative position labels to each neighbor node, the irregular neighbor size of each node makes it hard to construct the subgraph in parallel. Besides, due to the huge memory cost to store the adjacency matrix, for each end node pair, subgraphs are constructed with CPU-based neighbor samples. As a result, subgraph construction and encoding are not computationally efficient.
Secondly, \textbf{encoding with limited-sized neighborhood will lose information}.
Due to the low efficiency of neighbor sampling, most existing works only encode with a limited-sized subset of the edge high-order neighborhoods or only the first-order neighbors.  As a result, this approach fails to capture the necessary information needed for accurate modeling of the graph structure, which leads to relatively poor performance.
Thirdly, \textbf{extracting subgraph with single strategy ignores the diverse temporal information of different time intervals}. 
Existing works only encode with a single anchor-induced subgraph due to high computation cost. However, a temporal triangle of a recently interacted neighbor provides different information than a temporal triangle of a historically interacted neighbor. Therefore, single subgraph extraction ignores the diverse temporal information of different time intervals.

To address the issues above, we propose a novel structure encoding method named \textbf{C}o-\textbf{N}eighbor \textbf{E}ncoding \textbf{S}chema (\textbf{CNES}). 
% Unlike existing methods that encode with an irregular, high-order query-induced subgraph by CPU-based neighbor sampling and relative distance assignment, CNES generates structure encoding with regular, low-order subgraphs with memory indexing and vector-based parallel co-neighbor computation on GPU.  
% We assign a hashtable-based memory to each node in the temporal graph, the memory acts as a compressed adjacent matrix that can be stored on GPU, which replaces CPU-based neighbor sampling with GPU-based searching to construct a subgraph. 
% Correspondingly, CNES generates structure encoding for the edge-anchored subgraph by determining the common neighbors between the end node and the neighbor nodes through regular-sized vector computation in parallel for all node pairs in a batch to take the place of the relative position label assignment. 
% With the help of efficient subgraph construction and structure encoding computation, CNES can encode structure with a larger subgraph than existing sample-based methods while keeping high efficiency, resulting in a better performance. 
% To sum up, This method helps reduce the information loss in structure encoding caused by the conflict between the large number of neighbors and the limited subgraph size. 
It uses regular, low-order subgraphs with index querying and vector-based parallel co-neighbor computation on GPU to generate structure encoding. Each node in the temporal graph is assigned a hashtable-based memory, acting as a compressed adjacent matrix that can be stored on the GPU. In this way, we can replace CPU-based neighbor sampling with GPU-based index querying to access high-order structures. With hashtable-based memory, CNES determines the common neighbors between the other end node and the neighbor nodes through regular-sized vector computation in parallel for all node pairs in a batch, instead of assigning relative position labels, for the neighbor nodes' irregular high-order query-induced subgraphs. Unlike existing sample-based methods, CNES can encode structures with a larger neighborhood size while maintaining high efficiency, and reducing information loss in structure encoding caused by the limited subgraph size.
In addition, CNES introduces a \textbf{Temporal-Diverse Memory} to generate structure encoding for subgraphs at different time intervals. 

We also propose a dynamic graph learning framework named \textbf{C}o-\textbf{N}eighbor \textbf{E}ncoding \textbf{N}etwork (\textbf{CNE-N}), a light-cost model for dynamic link prediction. The proposed method is proven to be effective and efficient both in theory (as shown in Table \ref{tab:compare method}) and experimentally.

%We divide the high-order irregular end nodes-queried subgraph into multiple low-order regular neighbor-queried subgraphs and encode them in parallel with vector computation.
% employs a set-based neighbor memory to calculate the number of common neighbors between two query nodes for generating structure encoding. This method avoids the need for extracting expensive node high-order structural roles. Unlike other techniques that encode subgraphs in a single aspect, CNES generates multi-aspect structure encoding by treating each pair of nodes in the subgraph as query nodes. This approach accurately models the structural features without requiring high-order neighbor extraction, making it more efficient and suitable for parallel computation on a GPU. Additionally, CNES introduces a Long-Short Memory algorithm to generate structure encoding for subgraphs at different time intervals. The model's performance can be significantly improved by incorporating multi-aspect encoding in anchor nodes and time intervals.

Our contributions can be summarized as follows:

\begin{itemize}
    \item An efficient structure encoding method CNES is proposed. The method accelerates structure encoding by compressing the adjacency matrix to hashtable-based memory and computing co-neighbors with vector-based parallel computation. 
    %The encoding method can be easily extended to previous models.
    \item This paper generates a long-term and a short-term structure encoding for neighbors with temporal-diverse memory, enabling the capture of temporal structural patterns.
    %\item long-short memory with different storage dimensions is designed to generate structure encoding for subgraph different time intervals.
    \item Extensive experiments on 13 datasets show the effectiveness and the efficiency of the proposed method. Previous models equipped with the proposed structure encoding also result in a better performance.
\end{itemize}

%The method first extracts the first-order neighbors without expensive high-order neighbor sampling. Next, we generate structure encoding by treating each node pair in the subgraph as the two anchor nodes, instead of encoding with the two ends nodes as anchor nodes in previous works, and compute the number of the common neighbors of the two anchor nodes with a set-based neighbor memory for each node, the Encoding can be easily computed in parallel on GPU. Finally, Long-Short Memory is proposed to generate temporal subgraphs in different time intervals. 
%The rest of this paper is organized as follows: \secref{section-2} formalizes the studied problem. \secref{section-3} introduces the framework and each component of our approach. \secref{section-4} evaluates the proposed method by experiments. \secref{section-5} reviews the related research. \secref{section-6} concludes the entire paper.

\section{Preliminaries}
\label{section-2}

% This section presents the necessary definitions as well as the formalization of the studied problem. 

% \subsection{Definitions}
\begin{definition}
    \textbf{Dynamic Graph}. We define the dynamic graph as a sequence of edges $\mathcal{G}_T$ where $\mathcal{G}_T=\left\{e_1,\cdots,e_i,\cdots,e_n\right\}$, $0 \textless t_1 \leq \cdots \leq t_i \leq \cdots \leq t_n \leq T$ and $e_i=\left\{u,v,t\right\}$, $u$ and $v \in N$ represent the source node and the destination node of an edge, and $t$ means the edge occurs at timestamp $t$. $N$ is the node set of the dynamic graph. Each node $u \in N$ has a feature vector $x_N(u) \in R^{d_N}$, and each edge $(u, v, t)$ has an edge feature vector $x^t_E(u,v) \in R^{d_E}$, where $d_N$ and $d_E$ are the dimensions of the node feature and link feature. 
\end{definition}

\begin{definition}
    \textbf{k-order edge neighborhood subgraph}. We define the k-order edge neighborhood subgraph $S^t_k(u,v)$ as the neighbors within $k$ hops $N^k_u$ and $N^k_v$ of the two end nodes $u$ and $v$ at the timestamp $t$. The subgraph contains the historical interactions between the nodes.
\end{definition}

\begin{definition}
    \textbf{Problem Formalization}. Given a dynamic graph, $\mathcal{G}_t$, Dynamic Link Prediction (DLP) task aims to learn the dynamic representation $h^t_u$ and $h^t_v$ for the source node $u$ and destination node $v$ with the k-order edge neighborhood subgraph $S^t_k(u,v)$ and predict the interact probability $p^t_{u,v}$ of the two end nodes at timestamp $t$.
\end{definition}

\subsection{Structure encoding}

\begin{table}[htbp]
\caption{Comparing of different designs of structure encoding functions.}
\label{tab:SE method}
\resizebox{\linewidth}{!}{
\centering
\begin{tabular}{c|ccc}
\hline
\textbf{Method}    & \textbf{Compress Function}         & \textbf{Updation}     & \textbf{Involve Nodes} \\ \hline
\textbf{CAWN}      & time-weighted sample & recompute    & u,v           \\
\textbf{DyGformer} & recent sample       & recompute    & u,v           \\
\textbf{PINT}      & -    & u,v          & u,v,$N_u$,$N_v$  \\
\textbf{NAT}       & single hash function         & u,v          & u,v           \\
\textbf{CNEN (ours)}      & multiple hash function         & u,v,$N_u$,$N_v$ & u,v,$N_u$,$N_v$  \\ \hline
\end{tabular}
}
\vspace{-0.3cm}
\end{table}

\begin{figure}[htbp]
\centering
\includegraphics[width=0.9\linewidth]{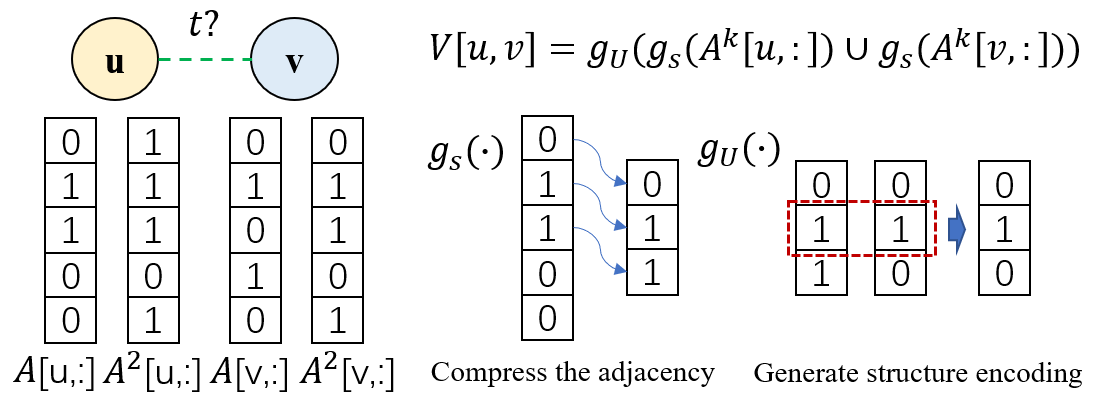}
\caption{Structure Encoding can be formalized as two functions, a compress function to get the neighbor set from the adjacency matrix, and a relation function to generate encoding from the union of the two neighbor sets. }
\label{Fig:structure encoding}
\vspace{-0.2cm}
\end{figure}

The structure encoding of the dynamic graph is to generate a structure feature $x^t_S(u, v) \in R^{d_S}$ for each edge $(u, v, t) \in \mathcal{G}_T$, where $d_S$ is the dimension of the structure feature. The structure encoding $V^t_k(u,v)$ represents the structural role of the source node $u$, the destination node $v$, and their neighbor nodes $\{i \in N^t_u \cup N^t_v\}$ in the $k$-order edge neighborhood subgraph.

As shown in Figure \ref{Fig:structure encoding}, we formalize the structure encoding as a compress function $g_n(\cdot)$ to get the neighbor set from the adjacency matrix, and a relation function $g_s(\cdot)$ to generate encoding from the union of the two neighbor sets. We then give a simpler mathematical form of structure encoding as follows:
\begin{gather}
    V^t_k(u,v) = \frac{1}{M}\sqrt{\sum^M_{i=0}(\hat{A^t_k}[u,i] \cdot \hat{A^t_k}[v,i])^2},
\end{gather}

where $\hat{A^t_k} = g_s(A^t_k) =  A^t_k \times \mathcal{M}^t_k$ is the compressed adjacency matrix, $\mathcal{M}^t_k \in R^{N \times M}$ is the compress matrix. We further summarize the difference in the structure encoding methods in Table \ref{tab:SE method}. 

The existing methods face a trade-off between accuracy and complexity. For example, CAWN and DyGformer achieve accuracy by recomputing the neighborhood with every update, but this process takes a lot of time. PINT stores the encoding without compression, requiring large storage space. NAT compresses the neighborhood using a single hash function and only updates the corresponding two center nodes when a link occurs, which means it cannot generate accurate up-to-date neighborhood encodings. 

Our method extends existing methods by balancing this trade-off. We follow NAT's approach to compress the matrix with a hash function and propose temporal diversity encoding to avoid information loss in hash conflicts. Additionally, we generate encoding and update memory with the k-order edge neighborhood subgraph instead of only the two nodes. As a result, CNEN achieves a balance between accuracy and complexity.

\subsection{Discussion of the Model Complexity} 
\label{sec:complexity}

According to Table \ref{tab:compare method}, we analyze the time and space complexity of each model under the following assumptions: 1) each model shares the same feature encoder, with hidden dimension $d = 50$; 2) subgraph-based model samples the same length $L = 20$ of sequence for every order; 3) Memory-based model shares a same sized memory with dimension $M = 100$, no matter what the memory stores, we assume the node feature size is also $M$ (because memory can be regarded as an aggregation of neighbor features, this assumption can help compare between models with/without memory); 4) The dynamic graph contains $N = 1000$ nodes.

\textbf{JODIE}\cite{kumar2019predicting} is a memory-based model. Therefore, the space complexity to store the node memories is $\mathcal{O}(NM)$; the model does not depend on the neighbor sample and computes link probability only with end nodes. Therefore, the time complexity to predict one link is $\mathcal{O}(M)+\mathcal{O}(Md)+\mathcal{O}(M) = \mathcal{O}(Md)$, where the two $\mathcal{O}(M)$ corresponding to memory readout and update, $\mathcal{O}(Md)$ is the complexity to encode the link features. The model is the most efficient baseline in theory and practice but may suffer from inductive issues (memory can not be generated for unseen nodes), and the model is not equipped with structure encoding.

\textbf{TGN}\cite{DBLP:journals/corr/abs-2006-10637} is a memory-based equipped with a graph neural network to encode high-order temporal information. It shares the same memory design as JODIE does. Therefore, the space complexity to store the node memories is also $\mathcal{O}(NM)$, the model extracts a $k$ order subgraph to gather information from neighbors; in practice, $k=1$, so the time complexity to predict one link is $L^{k=1}+L^{k=1} \times (\mathcal{O}(M)+\mathcal{O}(Md)+\mathcal{O}(M)) = \mathcal{O}(L+LMd)$. By neighbor sampling, the model overcomes inductive issues, and the model is not equipped with structure encoding.

\textbf{CAWN}\cite{xu2020inductive} is a structure encoding model with the random-walk neighbor sampler and does not depend on memory. The model extracts several numbers of walk sequences (we assume it to be 1) with sequence length $k$(for most datasets, $k \geq 2$). For each node in the sequence, the model assigns a relative position feature between it and the walk head node; the sequence is later encoded with an RNN-based model. Therefore the to predict one link is $L^{k=2}+L^{k=2} \times (\mathcal{O}(Md)) = \mathcal{O}(L^2+L^2Md)$. The model is computationally inefficient due to the complex and irregular subgraph construction.

\textbf{NAT}\cite{luo2022neighborhood} is a memory-based model but replaces the single vector-based memory in \textbf{JODIE} with dictionary-based memory. The representation of each neighbor is stored in the corresponding position without aggregation. Each memory contains $M_1$ position for first-order neighbors and $M_2$ position for second-order neighbors. For each neighbor, the model stores a representation of size $F$, the method keeps $(M_1 + M_2) \times F \approx M$, and we assume $M_1 + M_2 = L$ (for example, the default setting of the model is $M_1 = 32, M_2 = 16, F=4$). Therefore, the model does not depend on a neighbor sample to construct a subgraph and compute the structure encoding. As a result, the space complexity to store the node memories is $\mathcal{O}((M_1 + M_2) \times F) = \mathcal{O}(NM)$; the time complexity to predict one link is $\mathcal{O}((M_1 + M_2)Md) +\mathcal{O}(M_1d + M_2) = \mathcal{O}(LMd)$, where $\mathcal{O}((M_1 + M_2)Md)$ is the feature encoding for the $M_1 + M_2$ neighbors with feature dimension $M$. The model extends \textbf{JODIE} with structure encoding but still suffers from inductive issues.

\textbf{DyGFormer}\cite{yu2023towards} is a sequence-based model with first-order neighbor extraction. The model computes neighbor co-occurrence encoding by validating how often a neighbor node occurs in the neighbor sequences of the two end nodes. Therefore, the time complexity for structure encoding is $\mathcal{O}(L\times L)$ since it will compare over the whole sequence for each node. The encoding is expensive when $L$ is large. DyGFormer extracts a longer historical interaction sequence than other baseline models and reduces the time complexity with the patch technique (assume the patch size is 16). When encoding sequence with a Transformer encoder, the time complexity is $\mathcal{O}(L^2Md)$ before using the patch technique and is $\mathcal{O}((L/16)^2(16M)d) = \mathcal{O}(L^2Md/16)$.

\textbf{PINT}\cite{PINT2022} extends \textbf{TGN} with a link-wise relative position matrix $\in R^{N \times N}$ to store structure encoding between every pair of nodes, for each node, the model readout all corresponding relative positions in the matrix. As a result, the space complexity to store the node memories is $\mathcal{O}(N^2)$; the model extracts neighbor nodes in 2 2-order subgraphs. Therefore, the time complexity to predict one link is $L^{k=2}+L^{k=2} \times (\mathcal{O}(N) + \mathcal{O}(Nd) + \mathcal{O}(N)) = \mathcal{O}(L^2+L^2Nd)$, where the two $\mathcal{O}(N)$ corresponding to memory readout and update, $\mathcal{O}(Nd)$ is the complexity to encode the link features.

\textbf{CNE-N} replace the representation-based node memory in \textbf{TGN} to the neighbor set memory estimated by the hashtable, and the computation of co-neighbor encoding is regular with time complexity $\mathcal{O}(M)$. Therefore, the model shares the same complexity with \textbf{TGN} with $k=1$ ($M_L + M_S = M$). Where the space complexity to store the node memories is $\mathcal{O}(NM)$; the time complexity to predict one link is $L+L \times (\mathcal{O}(M)+\mathcal{O}(Md)+\mathcal{O}(M)) = \mathcal{O}(L+LMd)$.  

\section{Methodology}
\label{section-3}
\begin{figure}[htbp]
\centering
\includegraphics[width=0.9\linewidth]{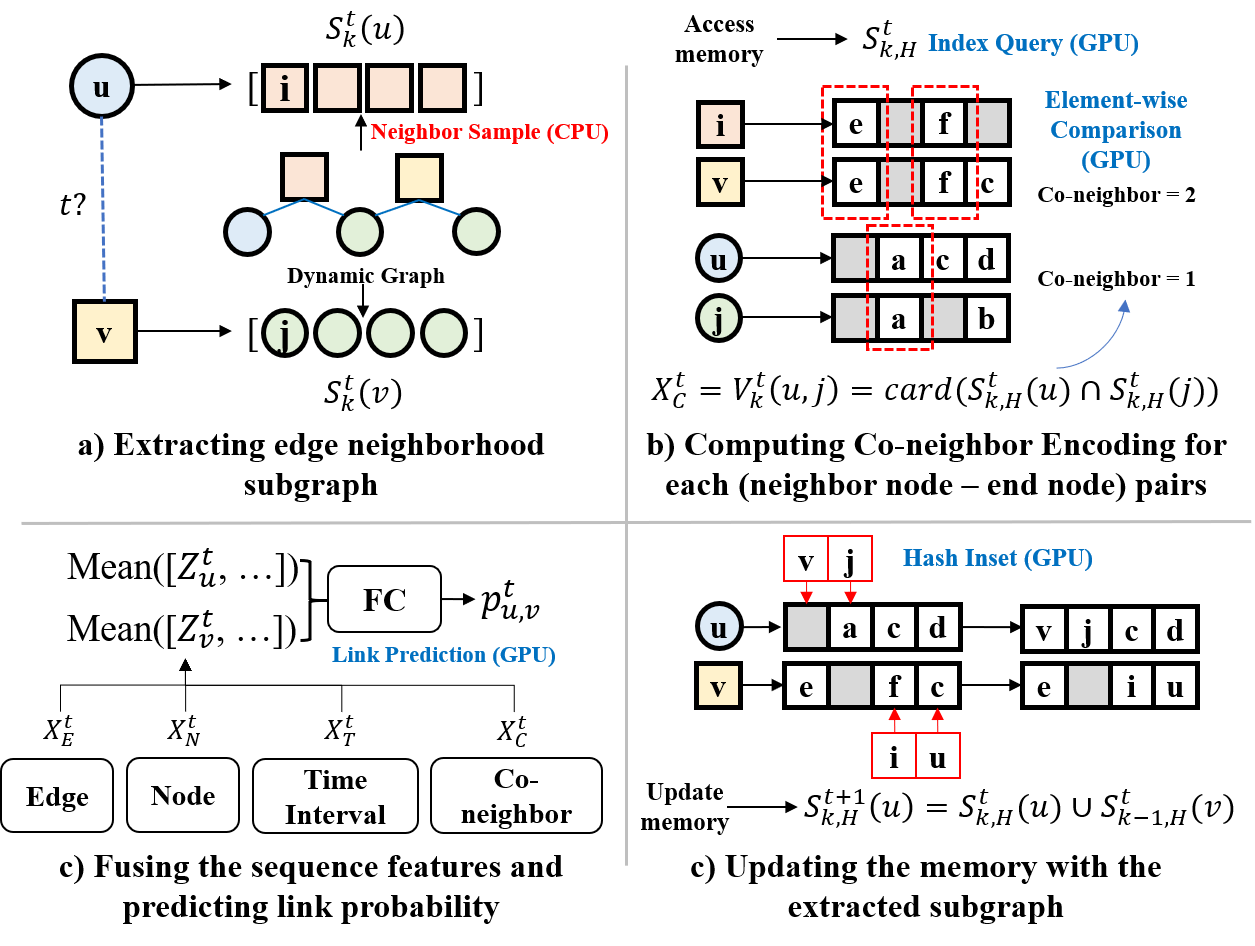}
\caption{We develop a dynamic graph learning framework Co-Neighbor Encoding Network (CNE-N), a light-cost model for dynamic link prediction. The model generates co-neighbor encoding with a hashtable-based memory and low-order neighborhood subgraph extracting, and the memory updates by the graph evolves. }
\label{Fig:framework}
\vspace{-0.4cm}
\end{figure}

This section presents the details of the Co-Neighbor Encoding Network(CNE-N). The framework of CNE-N is shown in Figure \ref{Fig:framework}. 
% \begin{figure*}[t!]
% \centering
% \includegraphics[width=1.0\linewidth]{pics/surel_example.png}
% \caption{We developed a corresponding framework Set-based Co-Neighbor Encoding Network (CNE-N), which is a light-cost and effective dynamic learning model for dynamic link prediction. The model generates co-neighbor encoding with a set-based neighbor memory.}
% \label{Fig:compare sequence}
% \end{figure*}

\subsection{Memory for structure Encoding}
We first introduce the hashtable-based memory in CNE-N to generate and update co-neighbor encoding, and long-short memory to generate structure encoding in different time intervals. All the computation can be done on GPU in parallel for multiple nodes in a batch.

\textbf{Co-neighbor Encoding.} 
Let $V^t_{k}(u,i)$ be the common neighbors for the two nodes $u$ and $i$ in the $k$-order edge neighborhood subgraph at timestamp $t$. We compute the co-neighbor encoding $[V^t_{k}(u,i), V^t_{k}(v,i)]$ for each node $i$ in the subgraph $S^t_k(u, v, t)$ of the link $(u,v)$. The count can be computed efficiently by first getting the intersection of $S^t_k(u)$ and $S^t_k(i)$ (where $S^t_k(i)$ corresponding to the node set of node $i$'s k-order subgraph, a node's 0-order subgraph is itself), and computing the cardinal number of it:

\begin{gather}
    V^t_{k}(u,i) = |S^t_k(u) \cap S^t_k(i)|, \\
    V^t_{k}(v,i) = |S^t_k(v) \cap S^t_k(i)|.
\end{gather}

\textbf{Hashtable-based Memory.}
In practice, we implement subgraph node sets with hash tables. Such data structure can estimate a set with a fixed-size sequence $H \in R^{|N| \times M}$, node-ids are mapped to the row index with a hash function $hash(i) = (q * i)(\textbf{mod} \ M)$ where $M$ is the size of the hash table $H^t$. We store node $a$ in $i$'s hash table $H^t_i$ by setting $H^t_i[hash(a)] = a$.

\textbf{Encoding Computation.} 
With hashtable $H^t_u, H^t_v$ and $H^t_i$, we can efficiently compute the co-neighbor encoding $[ V^t_{k}(u,i), V^t_{k}(v,i)]$ by counting the same element at the corresponding location:
\begin{gather}
\label{equ:co-neigbor_encoding}
    V^t_{k}(u,i) = \sum^{M-1}_{m = 0} (H^t_u[m] \ = \  H^t_i[m]),\\
    V^t_{k}(v,i) = \sum^{M-1}_{m=0} (H^t_v[m] \ = \  H^t_i[m]).
\end{gather}

\textbf{Memory Update.} 
A dynamic graph keeps evolving; the neighbors of a node are also time-varied, so whenever a new link occurs, the neighbor set memory of the nodes in the link-induced subgraph should be updated. The updating can be estimated by replacing the neighbor set memory with the union of the memory of the node and the updated neighbor $S^{t+1}_{k}(u) = S^t_{k}(u,t) \cup S^t_{k-1}(v,t)$, where $S^t_{k-1}(v)$ stands for the $k-1$-hop subgraph node-set, the nodes have become node $u$'s new updated neighbor in the $k$-hop subgraph. In practice, the update via set union can be implemented by hash\_insert $H^t_u[hash(v)] \gets v$.

\textbf{Temporal-Diverse Memory.} 
Hashtable-based memory stores all the neighbors of a node without temporal information from different intervals. When encoding a dynamic graph with a few nodes like Social Evo., the neighbor storage of each node will eventually be almost the same. Co-neighbor encoding can no longer provide useful structural information for link prediction. Besides, a single subgraph to encode the structure encoding can not use the evolving pattern in the dynamic structure, namely the temporal issue. Therefore, we propose long-short memory, which contains two parts of the hashtable: a short memory $H^s \in R^{|N| \times M_s}$ with a smaller hashtable size to store recent neighbor nodes and a long memory $H^l \in R^{|N| \times M_l}$ with a larger hashtable size to store more historical neighbor nodes. The two hashtables individually generate structure information and update the memory by the sequence. With the limited storage space, the short memory will replace neighbors in the hashtable more frequently. Therefore, it can generate co-neighbor encoding of the link short-term subgraph. We verify the effectiveness of temporal-diverse memory and report the result in Section \ref{sec:ablation}.

\subsection{Neural Encoding for Subgraph}
\label{sec:triangle}
In this section, we introduce how to predict future links with the proposed model CNE-N.

\textbf{Extracting edge neighborhood subgraph.}
We first extract the neighbors of the two end nodes to construct the $k$-order edge neighborhood subgraph with CPU-based neighbor sampling. In practice, to balance the trade-off between effectiveness and efficiency, we follow most existing works to extract the first-hop subgraph as historically interacted neighbors sequence with neighbor sampling, namely $S^t_1(u,v)$, 
given an interaction $(u,v,t)$, for the source node $u$ and the destination node $v$, we obtain the subgraphs that involves historical interactions of $u$ and $v$ before timestamp $t$ (including themselves), which are denoted by $S^t_1(u) = \{(u,i,\hat{t})|\hat{t} < t, i \in N^1_u\}$ and $S^t_1(v) = \{(v,i,\hat{t})|\hat{t} < t, i \in N^1_v\}$. In practice, to avoid the influence of the different lengths of historical interaction sequences, we pad each sequence with a fixed size $l_s$. As a result, $|S^t_1(u)| = |S^t_1(v)| = l_s$.

\textbf{Constructing Structure Encoding with Co-neighbor Encoding Schema.} With the proposed hashtable-based memory, we can efficiently compute the co-neighbor encoding between neighbor nodes in the source node $u$ neighborhood and the destination node $v$ neighborhood. Mathematically, given sequence $S^t_1(u)$ and $S^t_1(u)$, we can reach the stored neighbor memory of each corresponding node in the sequence as $S^t_{1, H^*}(u) \in R^{l_s \times M}$ and $S^t_{1, H^*}(v)\in R^{l_s \times M}$, and the neighbor hash table for source node $u$ and destination node $v$ as $H^{*,t}_u$ and $H^{*,t}_v$, where $*$ corresponding to $l$ for long memory and $s$ for short memory. Then we compute the co-neighbor encoding of the neighbor list $X^t_{u,C,*} \in R^{l_s \times 2} = [V_k^{*,t}(u,i), V_k^{*,t}(v,i)] \{i \in S^t_k(u)\}$ and $X^t_{v,C,*} \in R^{l_s \times 2} = $ $[V_k^{*,t}(v,j), V_k^{*,t}(u,j)]  \{j \in S^t_k(v)\}$ with Equation \ref{equ:co-neigbor_encoding}. 
% can be computed as :
% \begin{gather}
%     \resizebox{.9\hsize}{!}{$X^t_{u,C,*} = [\sum^{M-1}_{m = 0} (H^{*,t}_u[m] \ \textit{equals} \  S_{1, H^*}(u, t)[m]), \sum^{M-1}_{m = 0} (H^{*,t}_v[m] \ \textit{equals} \   S_{1, H^*}(u, t)[m])]$},\\
%     \resizebox{.9\hsize}{!}{$X^t_{v,C,*} = [\sum^{M-1}_{m = 0} (H^{*,t}_v[m] \ \textit{equals} \  S_{1, H^*}(v, t)[m]), \sum^{M-1}_{m = 0} (H^{*,t}_u[m] \ \textit{equals} \   S_{1, H^*}(v, t)[m])]$},
% \end{gather}

% \begin{gather}
%     \textbf{h}_{snq}^{t_i}  = \begin{cases}
%                 1, & \text{if } (s,q_i,t_i,r_i,k_i) \in S_s^t \land q_i = q\\
%                 0 , & \text{else}
%                     \end{cases}\\ 
%     \textbf{h}_{snk}^{t_i}  = \begin{cases}
%                 1, & \text{if } (s,q_i,t_i,r_i,k_i) \in S_s^t \land k_i = k\\
%                 0 , & \text{else}
%                     \end{cases}\\ 
%     \textbf{h}_{qn}^{t_i}  = \begin{cases}
%                 1, & \text{if } (s_i,q,t_i,r_i,k) \in S_q^t \land s_i = s\\
%                 0 , & \text{else}
%                     \end{cases}
% \end{gather}

For example, if the historical node for $u$ and $v$ are $[u, a, a]$ and $[v, a, u]$, given $V_1(u, a) = 4, V_1(u, v) = 5$ and $V_1(v, a) = 1$, the structure encoding for $u$ and $v$ are $[(M,5), (4,1), (4,1)]$ and $[(M,5), (1,4),$\\$(5,M)]$, where $M$ is the size of the hashtable. 

The co-neighbor encoding scheme is light cost and can be easily integrated into dynamic graph learning methods for better results. For each node, the method generates structure encoding by counting triangles between up to $1+l_s+l_sM$ nodes in parallel, which is larger than most of the existing models. For example, the DyGFormer only considers first-hop neighbors with up to $1+l_s$ nodes, and CAWN considers $1+l_s+l^2_s$ nodes but with high sample complexity. However, due to the high frequency of multiple-time interaction in the temporal graph, the sample-based model often repeatedly computes structure encoding between the same node pairs, which makes them consider fewer triangles than CNES with a hashtable-based memory to store neighbors. We demonstrate its efficiency in Section \ref{section-5}.

\textbf{Encoding Side information and Time Intervals.} As discussed in Section \ref{section-2}, there are some dynamic graphs with side information like node feature or edge feature, we encode those features in node $u$'s historical interaction sequence as $X^t_{u,N} \in R^{l_s \times d_N}$ and $X^t_{u,E} \in R^{l_s \times d_E}$. As for the time interval between the link and historical intervals, we follow TGAT\cite{xu2020inductive} and encode the time intervals $\Delta \hat{t} = t - \hat{t}$ with Fourier Time encoding as $X^t_{u,T} \in R^{l_s \times d_T}$: 

\begin{equation}
    \resizebox{.9\hsize}{!}{$X^t_{u,T} = \sqrt{\frac{1}{d_T}}[cos(\Delta \hat{t} w_1), sin(\Delta \hat{t} w_2), ... , cos(\Delta \hat{t} w_{d_T-1}), sin(\Delta \hat{t} w_{d_T})]$},
\end{equation}
where $w_1, \cdot, w_{d_T} \in R^{1 \times 1}$ are trainable parameters to encode time interval.

\textbf{Fusing Features.} We first assign each feature to the same dimension $d$ with several one-layer full connected neural networks $f(\cdot)$, and concatenated them together:
\begin{gather}
    Z^t_{u} = f_N(X^t_{u,N}) \parallel f_E(X^t_{u,E}) \parallel f_T(X^t_{u,T}) \parallel f_{C,l}(X^t_{u,C,l}) \parallel f_{C,s}(X^t_{u,C,s}),\\
    Z^t_{v} = f_N(X^t_{v,N}) \parallel f_E(X^t_{v,E}) \parallel f_T(X^t_{v,T}) \parallel f_{C,l}(X^t_{v,C,l}) \parallel f_{C,s}(X^t_{u,C,s}),
\end{gather}

Then, we employ a feature fusion encoder to encode the historical interaction sequence for subgraph-based embedding. The Layer Normalization (LN) is employed after every block. The input is processed via the feature fusion encoder to fuse the information from different features:
\begin{gather}
    Z^{t, l}_{u} = LN(Z^{t, l-1}_{u}W^l_{fu} + b^l_{fu}),\\
    Z^{t, l}_{v} = LN(Z^{t, l-1}_{v}W^l_{fu} + b^l_{fu}),
\end{gather}

where$W^l_{fu} \in R^{4d \times 4d}, b^l_{fu} \in R^{4d}$ are trainable parameters at the $l$-th layer of the encoder  $f_{fu} = \{*\}W_{fu} + b_{fu}$. The output of the L-th layer is $Z_u^{t, L}$ and $Z_v^{t, L} \in R^{l_s \times 5d}$.

\textbf{Generating Node Representation.} The time-aware representations of node $u$ and $v$ at timestamp $t$, $h^t_u \in R^{d_{o}}$ and $h^t_v \in R^{d_{out}}$ are derived by taking the mean pooling of their related representations in $Z^{t, L}$ with an output layer $f_{o} = \{*\}W_{o} + b_{o}$:
\begin{gather}
    h^t_u = (\frac{1}{l_s} \sum (Z^{t, L}_u))W_{o} + b_{o},\\
    h^t_v = (\frac{1}{l_s} \sum (Z^{t, L}_v))W_{o} + b_{o},
\end{gather}
where $W_{o} \in R^{5d \times d_{o}}, b_{o} \in R^{d_{o}}$ are trainable parameters at the output layer.

\textbf{Predicting Link Probability.} 
The predicted probability of the link $h^t_{u, v} \in R^d_{o}$ are merged with a merge layer $f_{m}$ as:
\begin{equation}
    p^t_{u, v} = \sigma((h^t_u, h^t_v)W_{m} + b_{m}),
\end{equation}
where $\sigma$ is the sigmoid function, $W_{m} \in R^{2d_{o} \times 1}, b_{m} \in R^{1}$ are trainable parameters at the merge layer.
We train the model with the Binary Cross-entropy loss between the positive link and the negative sampled node pair $[u, neg]$ with the random sample strategy:

\begin{equation}
    loss = -(log(p^t_{u, v}) + log(1 - p^t_{u, neg})).
\end{equation}

\textbf{Updating Node Memories in the subgraph.} Finally, we update the node co-neighbor memory with the new link-induced subgraph. Firstly, we update $u$ and $v$'s memory with the 1-order subgraph $H^{*}_u[hash(v)] \gets v$, $H^{*}_v[hash(u)] \gets u$, where $*$ corresponding to $l$ for long memory and $s$ for short memory, and then update the memory with neighbor nodes in the 2-order subgraph $H^{*}_u[hash(j)] \gets j, \{j \in S^t_1(v)\}$, $H^{*}_v[hash(i)] \gets i, \{i \in S^t_1(u)\}$, and we update the memory of the neighbor nodes $H^{*}_i[hash(v)] \gets v, \{i \in S^t_1(u)\}$, $H^{*}_j[hash(u)] \gets u, \{j \in S^t_1(v)\}$. We verify the effectiveness of updating the neighbor node memory and updating with the 2-order subgraph update in Section \ref{sec:ablation}.
% \begin{equation}
%     H^t_u = hash\_insert(H^{t-1}_u, [v ; S^t_u]),
% \end{equation}

% \begin{equation}
%     H^t_v = hash\_insert(H^{t-1}_v, [u ; S^t_v]),
% \end{equation}

% \begin{equation}
%     H^t_i = hash\_insert(H^{t-1}_i, u),  i \in S^t_u,
% \end{equation}

% \begin{equation}
%     H^t_i = hash\_insert(H^{t-1}_i, v),  i \in S^t_v,
% \end{equation}

% \begin{gather}
%     H^{*}_u[hash(v)] \to v\\
%     % H^{*}_v[hash(u)] \to u,
% \end{gather}
% where $*$ corresponding to $l$ for long memory and $s$ for short memory, and then update the memory with neighbor nodes in the 2-order subgraph:
% \begin{gather}
%     H^{*}_u[hash(j)] \to j {j \in S^t_1(v)},\\
%     % H^{*}_v[hash(i)] \to i {i \in S^t_1(u)}.
% \end{gather}

% And we update the memory of the neighbor nodes:
% \begin{gather}    
%     H^{*}_i[hash(v)] \to v {i \in S^t_1(u)},\\
%     % H^{*}_j[hash(u)] \to u {j \in S^t_1(v)}.
% \end{gather}

% We verify the effectiveness of updating the neighbor node memory and updating with the 2-order subgraph update in Section \ref{sec:ablation}.                    

\section{Experiments}
\label{section-4}

\subsection{Experimental Settings}
In this section, we introduce the experimental settings to examine the effectiveness and efficiency of the proposed model.

\begin{table*}[!htbp]
\caption{AP/AUC-ROC for transductive and inductive dynamic link prediction.}
\vspace{-0.1cm}
\label{tab:t1}
\resizebox{\textwidth}{!}{
% \centering
\begin{tabular}{@{}c|c|ccccccccccc@{}}
\hline
Metrics                                         & Datasets    & JODIE        & DyRep        & TGAT         & TGN          & CAWN         & EdgeBank          & TCL     & GraphMixer   & NAT          & DyGFormer   & CNE-N \\ 
\hline
{\multirow{13}{*}{Trans-AP}} & Wikipedia   & 96.50 ± 0.14 & 94.86 ± 0.06 & 96.94 ± 0.06 & 98.45 ± 0.06 & 98.76 ± 0.03 & 90.37 ± 0.00 & 96.47 ± 0.16 & 97.25 ± 0.03 & 97.50 ± 0.04 & {\ul 99.03 ± 0.02} & \textbf{99.09 ± 0.04} \\
                      & Reddit      & 98.31 ± 0.14 & 98.22 ± 0.04 & 98.52 ± 0.02 & 98.63 ± 0.06   & 99.11 ± 0.01 & 94.86 ± 0.00 & 97.53 ± 0.02 & 97.31 ± 0.01 & 99.10 ± 0.21 & \textbf{99.22 ± 0.01} & \textbf{99.22 ± 0.01} \\
                      & MOOC        & 80.23 ± 2.44 & 81.97 ± 0.49 & 85.84 ± 0.15 & {\ul 89.15 ± 1.60}   & 80.15 ± 0.25 & 57.97 ± 0.00 & 82.38 ± 0.24 & 82.78 ± 0.15 & 87.21 ± 0.63 & 87.52 ± 0.49 & \textbf{94.18 ± 0.07}\\
                      & LastFM      & 70.85 ± 2.13 & 71.92 ± 2.21 & 73.42 ± 0.21 & 77.07 ± 3.97   & 86.99 ± 0.06 & 79.29 ± 0.00 & 67.27 ± 2.16 & 75.61 ± 0.24 & 88.57 ± 1.76 & {\ul 93.00 ± 0.12} & \textbf{93.55 ± 0.12} \\
                      & Enron       & 84.77 ± 0.30 & 82.38 ± 3.36 & 71.12 ± 0.97 & 86.53 ± 1.11   & 89.56 ± 0.09 & 83.53 ± 0.00 & 79.70 ± 0.71 & 82.25 ± 0.16 & 90.81 ± 0.31 & {\ul 92.47 ± 0.12} & \textbf{92.48 ± 0.10} \\
                      & Social Evo. & 89.89 ± 0.55 & 88.87 ± 0.30 & 93.16 ± 0.17 & 93.57 ± 0.17   & 84.96 ± 0.09 & 74.95 ± 0.00 & 93.13 ± 0.16 & 93.37 ± 0.07 & 91.23  ± 0.37 & \textbf{94.73 ± 0.01} & {\ul 94.60 ± 0.03}\\
                      & UCI         & 89.43 ± 1.09 & 65.14 ± 2.30 & 79.63 ± 0.70 & 92.34 ± 1.04   & 95.18 ± 0.06 & 76.20 ± 0.00 & 89.57 ± 1.63 & 93.55 ± 0.57 & 94.26 ±  0.37 & {\ul 95.79 ± 0.17} & \textbf{96.85 ± 0.08} \\
                      & Flights     & 95.60 ± 1.73 & 95.29 ± 0.72 & 94.03 ± 0.18 & 97.95 ± 0.14   & 98.51 ± 0.01 & 89.35 ± 0.00 & 91.23 ± 0.02 & 90.99 ± 0.05 & 97.66 ±  0.80 & {\ul 98.91 ± 0.01}  &  \textbf{98.92 ± 0.01} \\
                      & Can. Parl.  & 69.26 ± 0.31 & 66.54 ± 2.76 & 70.73 ± 0.72 & 70.88 ± 2.34   & 69.82 ± 2.34 & 64.55 ± 0.00 & 68.67 ± 2.67 & 77.04 ± 0.46  &  83.83 ± 1.20 & \textbf{97.36 ± 0.45}  &{\ul 86.09 ± 0.14} \\
                      & US Legis.   & 75.05 ± 1.52 & 75.34 ± 0.39 & 68.52 ± 3.16 & 75.99 ± 0.58   & 70.58 ± 0.48 & 58.39 ± 0.00 & 69.59 ± 0.48 & 67.74 ± 1.02  & {\ul 77.56 ± 0.21} & 71.11 ± 0.59 & \textbf{78.69 ± 0.12}\\
                      & UN Trade    & 64.94 ± 0.31 & 63.21 ± 0.93 & 61.47 ± 0.18 & 65.03 ± 1.37   & 65.39 ± 0.12 & 60.41 ± 0.00 & 62.21 ± 0.03 & 62.61 ± 0.27 & {\ul 72.32 ± 0.69} & 66.46 ± 1.29 & \textbf{76.92 ± 0.06}\\
                      & UN Vote     & 63.91 ± 0.81 & 62.81 ± 0.80 & 52.21 ± 0.98 & 65.72 ± 2.17   & 52.84 ± 0.10 & 58.49 ± 0.00 & 51.90 ± 0.30 & 52.11 ± 0.16  & \textbf{69.70 ± 0.49} & 55.55 ± 0.42 &  {\ul 66.40 ± 0.12} \\
                      & Contact     & 95.31 ± 1.33 & 95.98 ± 0.15 & 96.28 ± 0.09 & 96.89 ± 0.56   & 90.26 ± 0.28 & 92.58 ± 0.00 & 92.44 ± 0.12 & 91.92 ± 0.03  & 97.25  ±  0.33 & {\ul 98.29 ± 0.01} & \textbf{98.40 ± 0.02}\\
\cline{2-13}
                      & Avg. Rank   &       7.08       &       7.85       &      7.62        &     4.38         &       5.92       &      9.61        &      8.76        &       7.77       &        3.31     &      {\ul 2.53}  &      \textbf{1.38}      \\ 
\hline
{\multirow{13}{*}{Trans-AUC}} & Wikipedia   & 96.33 ± 0.07 & 94.37 ± 0.09 & 96.67 ± 0.07 & 98.37 ± 0.07 & 98.54 ± 0.04 & 90.78 ± 0.00 & 95.84 ± 0.18 & 96.92 ± 0.03 &96.72 ± 0.21& {\ul 98.91 ± 0.02} &\textbf{99.01 ± 0.05} \\
                        & Reddit      & 98.31 ± 0.05 & 98.17 ± 0.05 & 98.47 ± 0.02 & 98.60 ± 0.06 & 99.01 ± 0.01 & 95.37 ± 0.00 & 97.42 ± 0.02 & 97.17 ± 0.02 & {99.02 ± 0.10} & {\ul 99.15 ± 0.01} &\textbf{99.15 ± 0.01} \\
                        & MOOC        & 83.81 ± 2.09 & 85.03 ± 0.58 & 87.11 ± 0.19 & 91.21 ± 1.15 & 80.38 ± 0.26 & 60.86 ± 0.00 & 83.12 ± 0.18 & 84.01 ± 0.17 & {\ul 88.38 ± 0.71} & 87.91 ± 0.58 & \textbf{95.69 ± 0.07} \\
                        & LastFM      & 70.49 ± 1.66 & 71.16 ± 1.89 & 71.59 ± 0.18 & 78.47 ± 2.94 & 85.92 ± 0.10 & 83.77 ± 0.00 & 64.06 ± 1.16 & 73.53 ± 0.12 & 86.94  ± 2.29 & {93.05 ± 0.10} & {\ul 93.13 ± 0.11}\\
                        & Enron       & 87.96 ± 0.52 & 84.89 ± 3.00 & 68.89 ± 1.10 & 88.32 ± 0.99 & 90.45 ± 0.14 & 87.05 ± 0.00 & 75.74 ± 0.72 & 84.38 ± 0.21 & 92.02 ± 0.32 & \textbf{93.33 ± 0.13} & {\ul 93.12 ± 0.09} \\
                        & Social Evo. & 92.05 ± 0.46 & 90.76 ± 0.21 & 94.76 ± 0.16 & 95.39 ± 0.17 & 87.34 ± 0.08 & 81.60 ± 0.00 & 94.84 ± 0.17 & 95.23 ± 0.07 & 93.22  ± 0.13 & \textbf{96.30 ± 0.01} & {\ul 96.24 ± 0.14} \\
                        & UCI         & 90.44 ± 0.49 & 68.77 ± 2.34 & 78.53 ± 0.74 & 92.03 ± 1.13 & 93.87 ± 0.08 & 77.30 ± 0.00 & 87.82 ± 1.36 & 92.52 ± 0.67 & 93.02 ±  0.48 & {\ul 94.49 ± 0.26} & \textbf{96.03 ± 0.08} \\
                        & Flights     & 96.21 ± 1.42 & 95.95 ± 0.62 & 94.13 ± 0.17 & 98.22 ± 0.13 & 98.45 ± 0.01 & 90.23 ± 0.00 & 91.21 ± 0.02 & 91.13 ± 0.01 & 97.32 ±  0.34 & {\ul 98.93 ± 0.01} & \textbf{98.96 ± 0.02} \\
                        & Can. Parl.  & 78.21 ± 0.23 & 73.35 ± 3.67 & 75.69 ± 0.78 & 76.99 ± 1.80 & 75.70 ± 3.27 & 64.14 ± 0.00 & 72.46 ± 3.23 & 83.17 ± 0.53 &  87.70 ± 1.37 & \textbf{97.76 ± 0.41} & {\ul 89.71 ± 0.07}  \\
                        & US Legis.   & 82.85 ± 1.07 & 82.28 ± 0.32 & 75.84 ± 1.99 & 83.34 ± 0.43 & 77.16 ± 0.39 & 62.57 ± 0.00 & 76.27 ± 0.63 & 76.96 ± 0.79 & \textbf{84.68 ± 0.35} & 77.90 ± 0.58 & {\ul 84.14 ± 0.63}\\
                        & UN Trade    & 69.62 ± 0.44 & 67.44 ± 0.83 & 64.01 ± 0.12 & 69.10 ± 1.67 & 68.54 ± 0.18 & 66.75 ± 0.00 & 64.72 ± 0.05 & 65.52 ± 0.51 & {\ul 76.76 ± 0.81} & 70.20 ± 1.44& \textbf{78.57 ± 0.07}\\
                        & UN Vote     & 68.53 ± 0.95 & 67.18 ± 1.04 & 52.83 ± 1.12 & 69.71 ± 2.65 & 53.09 ± 0.22 & 62.97 ± 0.00 & 51.88 ± 0.36 & 52.46 ± 0.27 & \textbf{74.44 ± 2.01} & 57.12 ± 0.62& {\ul 69.52 ± 0.34} \\
                        & Contact     & 96.66 ± 0.89 & 96.48 ± 0.14 & 96.95 ± 0.08 & 97.54 ± 0.35 & 89.99 ± 0.34 & 94.34 ± 0.00 & 94.15 ± 0.09 & 93.94 ± 0.02 & 97.64 ±  0.58 & {\ul 98.53 ± 0.01} & \textbf{98.77 ± 0.01} \\ 
\cline{2-13}
                      & Avg. Rank   &     6.31   &     7.69 &       7.92        &    4.23          &      6.15         &     9.31          &    9.23          &  7.69             &      3.15   & {\ul 2.85}    &   \textbf{1.31} 
    \\ 
% \hline

\hline
{\multirow{14}{*}{Ind-AP}}& Wikipedia   & 94.82 ± 0.20 & 92.43 ± 0.37 & 96.22 ± 0.07 & 97.83 ± 0.04 & 98.24 ± 0.03 & - & 96.22 ± 0.17 & 96.65 ± 0.02 & 95.40 ± 0.04 & \textbf{98.59 ± 0.03} & {\ul 98.37 ± 0.03} \\%10 64
                       & Reddit      & 96.50 ± 0.13 & 96.09 ± 0.11 & 97.09 ± 0.04 & 97.50 ± 0.07 & 98.62 ± 0.01 & - & 94.09 ± 0.07 & 95.26 ± 0.02 & 98.56 ± 0.21 & \textbf{98.84 ± 0.02} & \underline{98.78 ± 0.01} \\%64 128
                       & MOOC        & 79.63 ± 1.92 & 81.07 ± 0.44 & 85.50 ± 0.19 & {\ul 89.04 ± 1.17} & 81.42 ± 0.24 & - & 80.60 ± 0.22 & 81.41 ± 0.21 & 83.59 ± 1.58 & 86.96 ± 0.43 & \textbf{91.89 ± 0.31} \\%10 64
                       & LastFM      & 81.61 ± 3.82 & 83.02 ± 1.48 & 78.63 ± 0.31 & 81.45 ± 4.29 & 89.42 ± 0.07 & - & 73.53 ± 1.66 & 82.11 ± 0.42 & 86.87 ± 1.95 & {\ul 94.23 ± 0.09} & \textbf{94.64 ± 0.12} \\%100 64
                       & Enron       & 80.72 ± 1.39 & 74.55 ± 3.95 & 67.05 ± 1.51 & 77.94 ± 1.02 & 86.35 ± 0.51 & - & 76.14 ± 0.79 & 75.88 ± 0.48 &  89.03 ± 0.83 &  \textbf{89.76 ± 0.34} & {\ul 89.66 ± 0.22}\\%4 32
                       & Social Evo. & 91.96 ± 0.48 & 90.04 ± 0.47 & 91.41 ± 0.16 & 90.77 ± 0.86 & 79.94 ± 0.18 & - & 91.55 ± 0.09 & 91.86 ± 0.06 & 91.22 ± 0.32 & {\ul 93.14 ± 0.04} & \textbf{93.29 ± 0.37}\\%32 64
                       & UCI         & 79.86 ± 1.48 & 57.48 ± 1.87 & 79.54 ± 0.48 & 88.12 ± 2.05 & 92.73 ± 0.06 & - & 87.36 ± 2.03 & 91.19 ± 0.42 & 87.30 ±  0.15 & {94.54 ± 0.12} & \underline{95.03 ± 0.16} \\ %10 64
                       & Flights     & 94.74 ± 0.37 & 92.88 ± 0.73 & 88.73 ± 0.33 & 95.03 ± 0.60 & 97.06 ± 0.02 & - & 83.41 ± 0.07 & 83.03 ± 0.05 & 96.59 ± 1.67 & \textbf{97.79 ± 0.02} & {\ul 97.72 ± 0.04}\\%100 64
                       & Can. Parl.  & 53.92 ± 0.94 & 54.02 ± 0.76 & 55.18 ± 0.79 & 54.10 ± 0.93 & 55.80 ± 0.69 & - & 54.30 ± 0.66 & 55.91 ± 0.82 & 60.62 ± 2.06 & \textbf{87.74 ± 0.71} & \underline{68.31 ± 0.59} \\%10 64
                       & US Legis.   & 54.93 ± 2.29 & 57.28 ± 0.71 & 51.00 ± 3.11 & {\ul 58.63 ± 0.37} & 53.17 ± 1.20 & - & 52.59 ± 0.97 & 50.71 ± 0.76 & 57.54 ± 0.80 & 54.28 ± 2.87 & \textbf{59.44 ± 0.44}\\%10 64
                       & UN Trade    & 59.65 ± 0.77 & 57.02 ± 0.69 & 61.03 ± 0.18 & 58.31 ± 3.15 & 65.24 ± 0.21 & - & 62.21 ± 0.12 & 62.17 ± 0.31 & \textbf{69.29 ± 1.59} & 64.55 ± 0.62 & {\ul 66.58 ± 0.27}\\%20 64
                       & UN Vote     & 56.64 ± 0.96 & 54.62 ± 2.22 & 52.24 ± 1.46 & 58.85 ± 2.51 &  49.94 ± 0.45 & - &  51.60 ± 0.97 & 50.68 ± 0.44 & {\ul 66.35 ± 4.06} & 55.93 ± 0.39& \textbf{69.71 ± 0.48}\\%2 4
                       & Contact     & 94.34 ± 1.45 & 92.18 ± 0.41 & 95.87 ± 0.11 & 93.82 ± 0.99 & 89.55 ± 0.30 & - & 91.11 ± 0.12 & 90.59 ± 0.05 & 96.79 ± 0.37 & \textbf{98.03 ± 0.02} & \underline{98.02 ± 0.05}\\%10 64
\cline{2-13} 
		       & Avg. Rank   &7.23 & 	8.46 & 	7.62  &	6.00  &	6.08 & -  &	8.38  &	7.77  &	 3.62  &	{\ul2.69} 	 & \textbf{1.46} 
  \\
\hline
{\multirow{13}{*}{Ind-AUC}}& Wikipedia   & 94.33 ± 0.27 & 91.49 ± 0.45 & 95.90 ± 0.09 & 97.72 ± 0.03 & 98.03 ± 0.04 & - & 95.57 ± 0.20 & 96.30 ± 0.04 & 94.74 ± 0.44 & \textbf{98.48 ± 0.03} &{98.23 ± 0.01} \\
                      & Reddit      & 96.52 ± 0.13 & 96.05 ± 0.12 & 96.98 ± 0.04 & 97.39 ± 0.07 & 98.42 ± 0.02 & - & 93.80 ± 0.07 & 94.97 ± 0.05 & 97.99 ± 0.52 & \textbf{98.71 ± 0.01} &\underline{98.62 ± 0.01} \\
                      & MOOC        & 83.16 ± 1.30 & 84.03 ± 0.49 & 86.84 ± 0.17 & {\ul 91.24 ± 0.99} & 81.86 ± 0.25 & - & 81.43 ± 0.19 & 82.77 ± 0.24 & 6.13 ± 3.55 & 87.62 ± 0.51 &\textbf{92.76 ± 0.29} \\
                      & LastFM      & 81.13 ± 3.39 & 82.24 ± 1.51 & 76.99 ± 0.29 & 82.61 ± 3.15 & 87.82 ± 0.12 & - & 70.84 ± 0.85 & 80.37 ± 0.18 & 83.07 ± 2.32 & {\ul 94.08 ± 0.08} &\textbf{94.38 ± 0.08} \\
                      & Enron       & 81.96 ± 1.34 & 76.34 ± 4.20 & 64.63 ± 1.74 & 78.83 ± 1.11 & 87.02 ± 0.50 & - & 72.33 ± 0.99 & 76.51 ± 0.71 &  89.92 ± 0.72 & \textbf{90.69 ± 0.26} & {\ul 90.18 ± 0.15}\\
                      & Social Evo. & 93.70 ± 0.29 & 91.18 ± 0.49 & 93.41 ± 0.19 & 93.43 ± 0.59 & 84.73 ± 0.27 & - & 93.71 ± 0.18 & 94.78 ± 1.00 & 92.11 ± 0.07 & \textbf{95.29 ± 0.03} & {\ul 95.16 ± 0.14}\\
                      & UCI         & 78.80 ± 0.94 & 58.08 ± 1.81 & 77.64 ± 0.38 & 86.68 ± 2.29 & 90.40 ± 0.11 & - & 84.49 ± 1.82 & 89.30 ± 0.57 & 83.81 ±  1.28 & {\ul 92.63 ± 0.13} &\textbf{93.34 ± 0.17} \\
                      & Flights     & 95.21 ± 0.32 & 93.56 ± 0.70 & 88.64 ± 0.35 & 95.92 ± 0.43 & 96.86 ± 0.02 & - & 82.48 ± 0.01 & 82.27 ± 0.06 & 96.36 ± 1.51 & {\ul 97.80 ± 0.02} & \textbf{97.82 ± 0.04}\\
                      & Can. Parl.  & 53.81 ± 1.14 & 55.27 ± 0.49 & 56.51 ± 0.75 & 55.86 ± 0.75 & 58.83 ± 1.13 & - & 55.83 ± 1.07 & 58.32 ± 1.08 & 61.62 ± 2.50 & \textbf{89.33 ± 0.48} & \underline{70.22 ± 0.77}\\
                      & US Legis.   & 58.12 ± 2.35 & 61.07 ± 0.56 & 48.27 ± 3.50 & {\ul 62.38 ± 0.48} & 51.49 ± 1.13 & - & 50.43 ± 1.48 & 47.20 ± 0.89 & \textbf{62.85 ± 0.84} & 53.21 ± 3.04 & 61.52 ± 0.52\\
                      & UN Trade    & 62.28 ± 0.50 & 58.82 ± 0.98 & 62.72 ± 0.12 & 59.99 ± 3.50 & 67.05 ± 0.21 & - & 63.76 ± 0.07 & 63.48 ± 0.37 & \textbf{72.56 ± 1.47} & 67.25 ± 1.05 &{\ul 67.26 ± 0.10}\\
                      & UN Vote     & 58.13 ± 1.43 & 55.13 ± 3.46 & 51.83 ± 1.35 & 61.23 ± 2.71 & 48.34 ± 0.76 & - & 50.51 ± 1.05 & 50.04 ± 0.86 & \underline{66.26 ± 5.48} & 56.73 ± 0.69 &\textbf{69.17 ± 0.53}\\
                      & Contact     & 95.37 ± 0.92 & 91.89 ± 0.38 & 96.53 ± 0.10 & 94.84 ± 0.75 & 89.07 ± 0.34 & - & 93.05 ± 0.09 & 92.83 ± 0.05 & 96.67 ± 0.45& \underline{98.30 ± 0.02} &\textbf{98.38 ± 0.02}\\
\cline{2-13} 
		       & Avg. Rank   &     6.62         &    7.77    &        7.23     &          5.31     &       5.62       & -  &     7.92          &          7.15     &      3.31 &        {\ul 2.62}       &      \textbf{1.46} 
        \\
\hline
\end{tabular}
}
\vspace{-0.4cm}
\end{table*}

\textbf{Datasets}. We test CNE-N and baseline models on thirteen commonly used datasets: Wikipedia, Reddit, MOOC, LastFM, Enron, Social Evo., UCI, Flights, Can. Parl., US Legis., UN Trade, UN Vote, and Contact, which are collected by Edgebank\cite{poursafaeitowards}. Details of the datasets are shown in Appendix \ref{sec:dataset}.

\textbf{Evaluation Metrics}. We follow TGN\cite{DBLP:journals/corr/abs-2006-10637} to evaluate the model performance on the dynamic link prediction task, which is to predict whether two nodes interact with each other at a certain time. The task contains two settings: transductive setting predicts future links between previously observed nodes in the training process and inductive setting tests link prediction between unseen nodes. Average Precision (AP) and Area Under the Receiver Operating Characteristic Curve (AUC-ROC) are adopted as the evaluation metrics. We use the random negative sample strategy to generate negative links. We considered a 70\%-15\%-15\% (train-val-test) split for each dataset. 

\textbf{Baselines}. We compare CNE-N with ten popular continuous-time dynamic graph learning baselines, including two memory-based models, JODIE\cite{kumar2019predicting} and DyRep\cite{dyrepLearningRepresentationsDynamicGraphs}. Two graph convolution-based models, TGAT\cite{xu2020inductive} and TGN\cite{DBLP:journals/corr/abs-2006-10637}; four sequence-based models, EdgeBank\cite{poursafaeitowards}, TCL\cite{wang2021tcl}, GraphMixer\cite{cong2023we}, and DyGFormer\cite{yu2023towards}; and two structure encoding-based models, NAT and CAWN\cite{wang2021inductive}. Details of the baseline models are listed in Appendix \ref{sec:baseline}. We test our model and baselines under DyGLib\cite{yu2023towards}. The code is available in \url{https://github.com/ckpassenger/DyGLib_CNEN/tree/CNEN}.

\textbf{Model Configurations}. For baselines, we follow the result reported by DyGFormer\cite{yu2023towards}. As for CNE-N,  We set the size of the long memory hashtable $M_l$ to 64 and set the short memory size $M_s$ to 16. We search the historical interaction sequence with a maximum length $l_s$ in $[4, 10, 20, 32, 64, 100]$. For all layers in the model, we set their hidden dimension $d_T, d, d_{o}$, and $d_{m}$ to 50.

\textbf{Implementation Details}. We employ the Adam optimizer with a learning rate of 0.0001 in all our experiments. The batch size of the input data is fixed at 200 for training, validation, and testing. The dropout rate was set at 0.1 for all experiments and all datasets. We conduct the experiments five times and took the average of the results. The experiments are carried out on a Windows machine with an AMD RYZEN 7 5800 CPU @ 3.30GHz having 6 physical cores. The GPU device used is an NVIDIA GTX3060 with a memory capacity of 12 GB.

\subsection{Performance Comparison}
In this section, we compare the performance between our method and baselines.

\begin{figure}[t]
\centering
\includegraphics[width=1.0\linewidth]{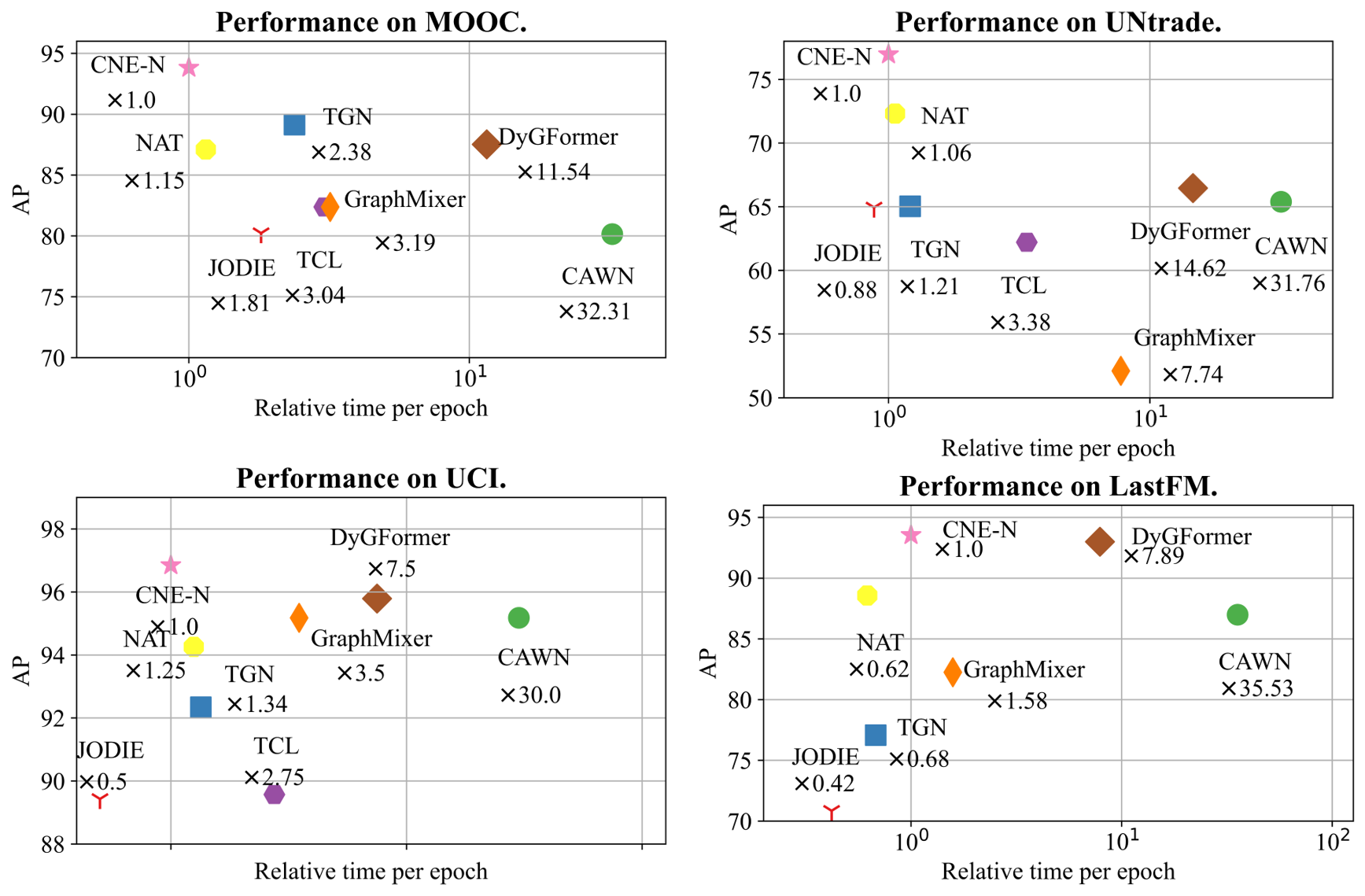}
\caption{CNE-N vs SOTA CTDG methods on MOOC, UNtrade, UCI, and LastFM. The horizontal axis shows the relative training time for each method as a multiple of CNE-N’s running time. The vertical axis shows average precision. 
}
\vspace{-0.6cm}
\label{Fig:compare speed}
\end{figure}

\textbf{Result}. We report the performance of different methods on the AP/AUC-ROC metric for the transductive setting of dynamic link prediction in Table \ref{tab:t1}, and in the inductive setting in \ref{tab:t2}. Note that EdgeBank\cite{poursafaeitowards} can be only evaluated for transductive dynamic link prediction, so its results under the inductive setting are not presented. For some of the results, we follow the results reported by their original papers. We also test model scalability in large-scale TGB dataset, please refer to Appendix \ref{sec:TGB} for more discussion. Furthermore, the performance and time per epoch for different dynamic learning methods are presented in Figure \ref{Fig:compare speed}. 
From Table \ref{tab:t1}, and Figure \ref{Fig:compare speed}, we have two main observations.

(1) In all datasets, except for Can. Parl. and Social Evo., CNE-N outperforms other baselines. The co-neighbor encoding performs best in social or interaction datasets with sufficiently large neighbors. This result confirms the effectiveness of our proposed co-neighbor encoding schema. The structure encoding helps the model differentiate between new links among highly related and unrelated nodes. However, Can. Parl is a policy dataset and highly relies on first-order neighbor modeling\cite{yu2023towards}. Therefore, DyGFormer performs well with long neighbor sequence. CNE-N fails to generate a precise encoding because the model depends on the number of common neighbors instead of individual neighbor information. Additionally, Social Evo. has very few nodes, leading to all hashtable-based memory being the same, rendering co-neighbor encoding useless. This inspired us to develop temporal-diverse memory to encode recent high-frequency interacting neighbors.

(2) CNE-N achieves the best performance with low computation cost. 
This is due to three reasons. Firstly, CNE-N constructs an edge neighborhood subgraph with a short first-order sequence length and considers high-order neighbors by accessing memory, which is more efficient than other sample-based subgraph construction methods like TGAT and TGN. Secondly, CNE-N computes co-neighbor encoding through vector-based parallel computing, which is more efficient than other structure encoding methods like CAWN or DyGFormer. Lastly, unlike other memory-based methods such as NAT and Jodie, CNE-N does not store the node's hidden state in memory and instead accesses/updates the memory using neural networks.

\subsection{Ablation Study}
\label{sec:ablation}
\begin{figure*}[!htbp]
\centering
\subfigure[transductive setting]
{
\begin{minipage}[b]{0.42\linewidth}
    \includegraphics[width=1.0\linewidth]{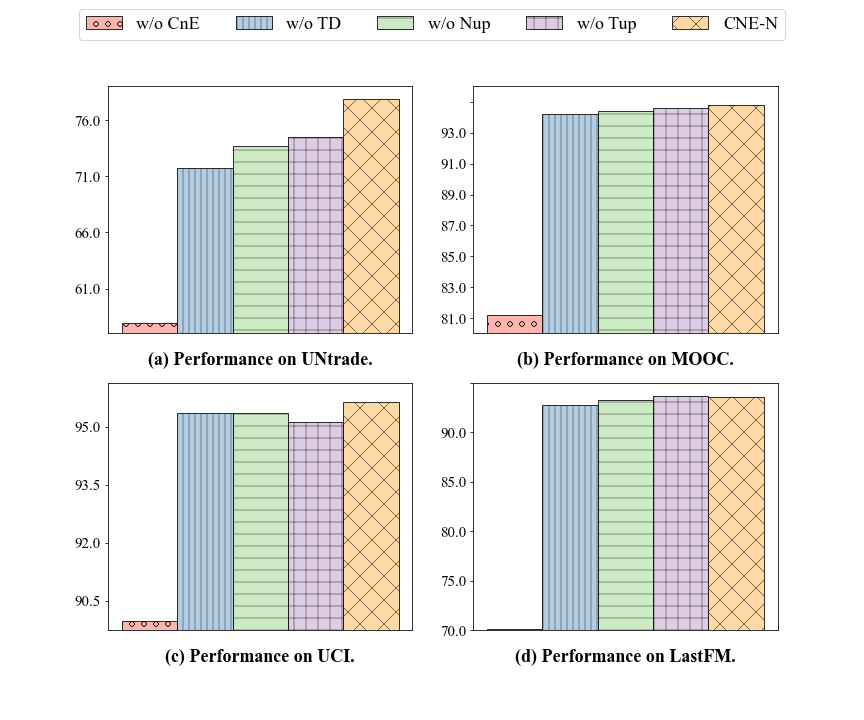}
\end{minipage}
}
\subfigure[inductive setting]
{
\begin{minipage}[b]{0.42\linewidth}
    \includegraphics[width=1.0\linewidth]{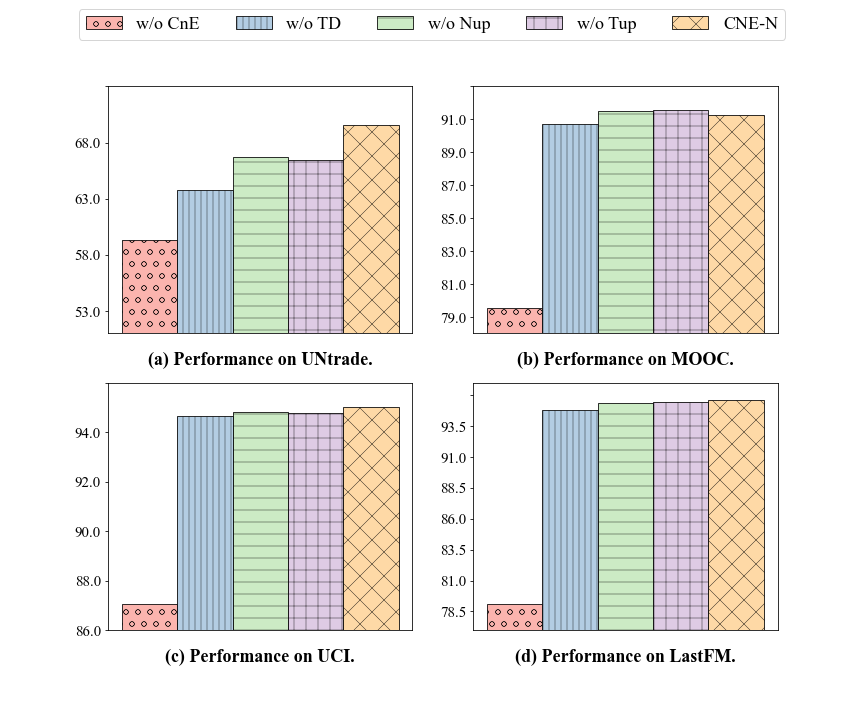}
\end{minipage}
}
\vspace{-0.4cm}
\caption{AP in both settings for ablation study of the CNE-N.}
\label{fig:ablation}
\vspace{-0.4cm}
\end{figure*}

We conducted an ablation study to validate the effectiveness of certain designs in CNE-N. This included examining the use of Co-neighbor Encoding (CnE), the use of temporal-diverse memory (TD), updating neighbor memory (Nup), and updating memory with neighbors in the two-order subgraph (Tup). We removed each module separately and referred to the remaining parts as w/o CnE, w/o TD, w/o Nup, and w/o Tup. We evaluated the performance of different variants on MOOC, LastFM, UCI, and UN Trade datasets from four domains and presented the results in Figure \ref{fig:ablation}. Our findings indicate that CNE-N mostly performs best when using all the components. Co-neighbor encoding has the most significant impact on performance as it effectively captures the structural encoding between query node pairs in the subgraph. The temporal-diverse memory benefits model performance by generating structure encoding for subgraph different time intervals. Updating neighbor memory can increase the influence of newly occurred links, providing more information to predict future links between higher-order neighbors. Updating memory with two-order neighbors achieved the best performance in some settings. For instance, in a bipartite dataset like LastFM, two-order neighbors may replace other neighbors in the hashtable as they cannot provide much helpful information, to be more specific, an user does not directly interact with another user.

%UNtrade CnE-LS-Nup-Tup 6646/6455-7905/6528-7344/5708-8208/6780-8311/6723
%MOOC CnE-LS-Nup-Tup 8710/8625-9783/9023-9780/9084-9791/9473-9837/9225
%UCI CnE-LS-Nup-Tup 9579/9454-9832/9746-9828/9708-9862/9750-9864/9780
%LastFM CnE-LS-Nup-Tup 7163/7850-9664/9283-9562/9211-9788/9666-9665/9559

\subsection{Parameter Sensitivity}

In this part, we test how the number of triangles being considered in the subgraph influences model performance. As discussed in Section \ref{sec:triangle}, the number of nodes being considered is up to $1+l_s+l_sM$, to support our motivation that a larger subgraph results in a better model performance, we test model performance by changing over the two parameters.

\begin{figure}[!htbp]
\centering
\includegraphics[width=0.92\linewidth]{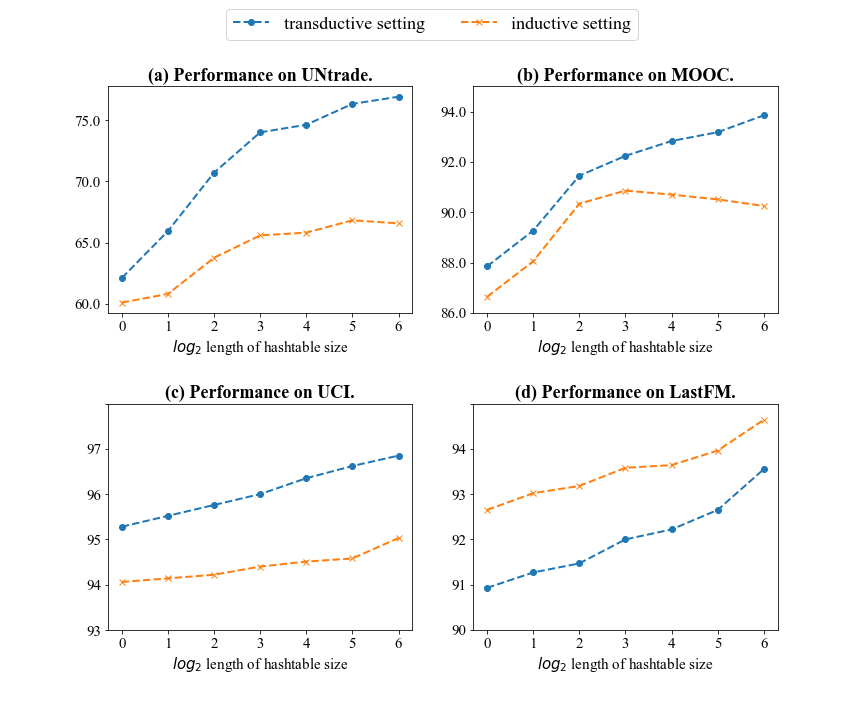}
\caption{AP in the two settings with varying hashtable size. }
\label{Fig:hashtable size}
\vspace{-0.3cm}
\end{figure}

\textbf{Sizes of Hashtables}. Hashtable size $M$ also influences the co-neighbor encoding accuracy. Reducing the size of the hashtable will increase the likelihood of hash conflict, which can affect the accuracy of co-neighbor encoding. We evaluated the sensitivity of the hashtable size parameter on MOOC, LastFM, UCI, and UN Trade, as shown in Figure \ref{Fig:hashtable size}. The results indicate that in the transductive setting, model performance improves as we increase the hashtable size. However, in the inductive setting, performance is negatively impacted when the size of the hashtable is too large.

\begin{figure}[!htbp]
\centering
\includegraphics[width=0.92\linewidth]{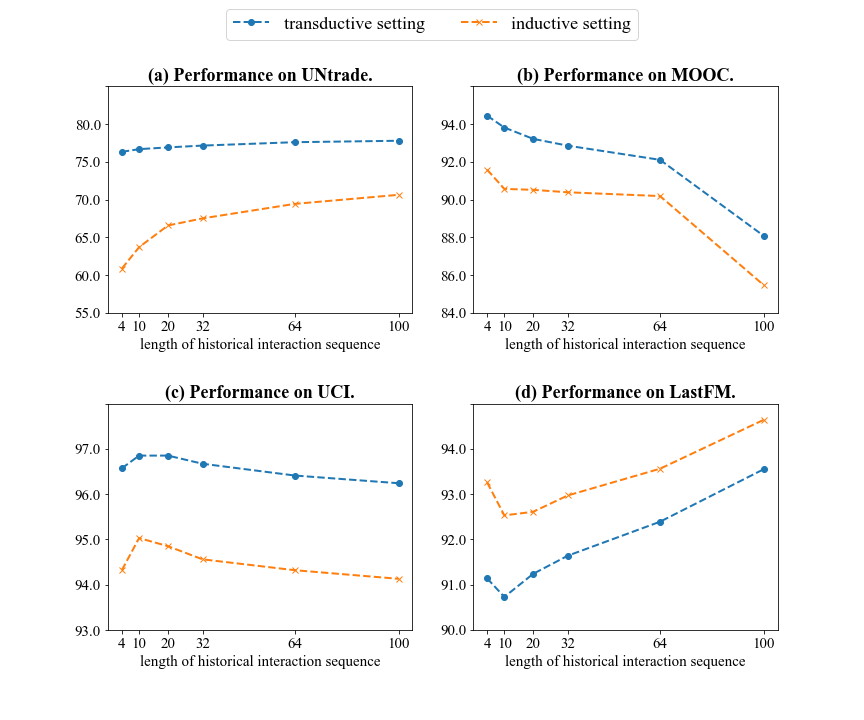}
\caption{AP in the two settings with varying historical interaction sequence length.}
\label{Fig:sequence length}
\vspace{-0.8cm}
\end{figure}

\textbf{Length of the Sequence}. The length of the historical interaction sequence $l_s$ is an important factor as it contains both valuable information from neighbors and noise and also affects the number of neighbor-induced subgraphs. In Figure \ref{Fig:sequence length}, we tested the sensitivity of the Sequence length parameter on MOOC, LastFM, UCI, and UN Trade datasets. For social datasets like UCI and MOOC, recent neighbors can provide valuable information, whereas former neighbors bring more noise. On the other hand, for user-item interaction datasets like UNtrade and LastFM, longer historical information can better model user preference.

\section{Related work}
\label{section-5}

% This section reviews the existing related literature and also points out their differences from our work.
% including academic network analysis, paper citation prediction, and graph representation learning. 
\subsection{Link Prediction in Graph}Previous work on temporal network representation learning used GNNs to process static graph snapshot sequences taken at time intervals, because of limitations in expressive power such as the inability to count triangles and the inability to distinguish nodes with the same structural roles, graph neural networks (GNNs) usually cannot extract the network evolution patterns, such as the triadic closure rule that is common in social networks, and perform poorly on link prediction (LP) tasks\cite{wang2021inductive,chamberlain2022graph}.
To enhance the expressivity of GNNs on LP tasks, many methods have been proposed, Assigning unique node IDs can distinguish different structural node representations, but at the expense of generalization  \cite{surprisingPowerGNNRandomNodeInit} and training convergence \cite{randomFeaturesStrengthenGNN}. However, these methods are still limited by the ability of GNNs to extract structural features from the shared neighborhood of multiple nodes.
To simplify computation and improve generalization ability, the state-of-the-art LP methods only perform computation on subgraphs containing links, given a set of queried node sets, CAW-N \cite{wang2021inductive}, SEAL\cite{linkPredictionBasedOnGNN, labelingTrickTheoryOfUsingGNNForMultiNodeRepLearning}, GraIL\cite{inductiveRelationPredictionBySubgraphReasoning} and SubGNN\cite{subgraphNeuralNetworks} and other SGRL models first extract subgraphs (named query-induced subgraphs) around the queried node sets, and then encode the extracted subgraphs for prediction \cite{chamberlain2022graph}. A large number of studies have shown that SGRL models are more robust[14], and more expressive\cite{understandingExtendingSubgraphGNNs, improvingGNNExpressivityViaSubgraphIsomorphismCounting}, compared with more complex techniques.

\subsection{Dynamic Graph Learning}
Representation learning on dynamic graphs is divided into DTDG /CTDG methods. Discrete-time dynamic graph learning (DTDG) methods, based on a low-cost discrete mechanism, tend to split event sequences into several individual snapshots\cite{liben2007link}. Each snapshot contains a part of the event sequence by time interval from one month to one year. The model treats each snapshot as a static graph with time features in edge and encodes them with traditional graph neural networks for spatial information propagation\cite{chen2022gc}. Some works factorized the connectivity matrix of each snapshot for node representation\cite{yu2017link, ma2019embedding, dunlavy2011temporal}. Another strategy is to take random walks on the snapshot and model the walk behaviors\cite{de2018combining,du2018dynamic}. Node embedding was later aggregated by weighted sum \cite{zhu2012hybrid}, temporal smoothness constraint, or sequential models like LSTM\cite{graves2012long} in E-LSTM-D\cite{chen2019lstm} and Know-Evolve\cite{trivedi2017know}. Those sequential models were added along the snapshots to capture the temporal evolving information among nodes in each snapshot. DTDG method can apply to tasks that have seasonal patterns since encoding along snapshots tends to learn patterns in a regular time interval. But at the same time, these methods require manual determination of time intervals, ignoring the temporal order of nodes in each snapshot, and have low accuracy.
%DTDG methods are also known as spatial-temporal networks. For example, DCRNN\cite{li2017diffusion} and STGCN\cite{yu2017spatio} apply the DTDG method on traffic area to model the traffic flow in different time intervals. 

Continuous-Time Dynamic Graph Learning (CTDG) methods are based on timestamped graphs\cite{trivedi2017knowevolve} (where evolution is represented as a continuous-time function), and typically have better flexibility and achieve higher accuracy. These methods split the event sequence into fixed-length edge batches (200 to 600 events per batch), leading to a large number of batches, therefore, to control the computation cost, CTDG methods only pass the message on a sampled local subgraph, mailbox (also known as memory) were designed to store information to be propagated between different subgraphs. Representative methods include temporal random walk\cite{wang2021inductive, luo2022neighborhood} and neural extensions of temporal point processes \cite{trivedi2017knowevolve, trivedi2019dyrep, zuo2018embedding}.  DynRep\cite{trivedi2019dyrep} and JODIE\cite{kumar2019predicting} developed a memory layer to store history edge messages and evolve node representation with a recurrent-based model when an edge occurs. 
TGAT\cite{xu2020inductive} sample on a temporal graph for a local subgraph, and generate node representation by graph embedding layer with random Fourier features to encode timestamps. 
TGN generalizes the graph network model of static graphs and most graph message passing type architectures\cite{rossi2020temporal}, which combines Jodie and TGAT.
Edgebank\cite{poursafaeitowards} discusses the influence of different negative sample strategies. Besides, Causal Anonymous Walk-Networks (CAW-N)\cite{wang2021inductive} computes the node embedding with a set of anonymized random walks, and aggregates them with recurrent model and attention operation.  

\section{Conclusion}
\label{section-6}
In this paper, we discussed the trade-off between accuracy and complexity in structure encoding in CTDG methods. We proposed a Co-Neighbor Encoding Schema (CNES) method to achieve the balance. CNES generates structure encoding by attaching a regular-sized hashtable-based neighbor memory to each node and calculating the number of common neighbors between the end node and each neighbor node with vector-based parallel computation. A Temporal-Diverse Memory is used to generate structure encoding for subgraphs at different time intervals. With the aforementioned techniques, we proposed a dynamic graph learning model called CNE-N. It achieves the highest performance in 10 of 13 datasets while being computationally efficient.

% \newpage

\section{Acknowledgments}
This work was supported by the National Natural Science Foundation of China (62272023, 51991395, 51991391, U1811463) and, the S\&T Program of Hebei(225A0802D).
\bibliographystyle{ACM-Reference-Format}
\bibliography{reference}

% \clearpage
\clearpage
\appendix
\section{Appendix} 
\label{section-appendix}
In the appendix, details of the experiments are introduced.
% \usepackage{booktabs}

\subsection{Descriptions of Datasets} 
\label{sec:dataset}
These thirteen datasets (Wikipedia, Reddit, MOOC, LastFM, Enron, Social Evo., UCI, Flights, Can. Parl., US Legis., UN Trade, UN Vote, and Contact) are collected by Edgebank\cite{poursafaeitowards} and cover diverse domains, detailed statistics are shown in Table \ref{tab:statistics}.\\

\textbf{Wikipedia:} consists of the bipartite interaction graph between editors and Wiki pages over a month \cite{poursafaeitowards}. Nodes represent editors and pages, and links denote the editing behaviors with timestamps. The features of the links are 172-dimensional Linguistic Inquiry and Word Count (LIWC) vectors.

\textbf{Reddit:} records thebipartite posts of users under subreddits during one month\cite{kumar2019predicting}. Users and subreddits are the nodes, and links are the timestamped posting requests. Each link has a 172-dimensional LIWC feature. 

\textbf{MOOC:} is a  a student bipartite interaction network of online sources units and models the student’s access to course videos or questions as links, and represents them by four-dimensional feature vectors\cite{kumar2019predicting}.

\textbf{LastFM:} is a bipartite network that records the songs listened to by users over one month. Users and songs are nodes, and links represent the listening behaviors of users.

\textbf{Enron:} records the email communications between employees of the ENRON energy corporation over three years. 

\textbf{UCI:} is an online communication network that models university students as nodes and messages posted by the students as links.

\textbf{Flights:} is a dynamic flight network that illustrates the evolution of air traffic during the COVID-19 pandemic. Airports are represented by nodes, and links denote tracked flights. Each link is associated with a weight indicating the number of flights between two airports in a day.

\textbf{Can. Parl.:} is a dynamic political network that captures interactions between Canadian Members of Parliament (MPs) from 2006 to 2019. Each node in the network represents an MP from an electoral district, and a link is established between two MPs when they both vote “yes” on a bill. The weight of each link corresponds to the number of times one MP voted “yes” for another MP in a year.

\textbf{US Legis.:} is a network that tracks the social interactions between legislators in the US Senate. It records the number of times two congresspersons have co-sponsored a bill in a given congress, and the weight of each link represents this number.

\textbf{UN Trade:} tracks the food and agriculture trade between 181 nations for over 30 years. The weight of each link represents the total sum of normalized agriculture import or export values between two specific countries.

\textbf{UNvote:} records roll-call votes in the United Nations General Assembly. If two nations both voted "yes" to an item, the weight of the link between them is increased by one. 

\textbf{Contact:} describes how the physical proximity evolves among about 700 university students over a month. Each student has a unique identifier and links denote that they are within 16 proximity to each other. Each link is associated with a weight, revealing the physical proximity between students. 

% \usepackage{booktabs}
\begin{table*}[t]
\caption{Statistics of the datasets}
\label{tab:statistics}
\resizebox{0.8\linewidth}{!}{
\begin{tabular}{@{}c|cccccc@{}}
\toprule
Datasets    & Domains     & \#Nodes & \#Links   & \#Node \& Link Features & Bipartite & Duration      \\ 
\hline
Wikipedia   & Social      & 9,227   & 157,474   &         - \& 172         & True      & 1 month       \\
Reddit      & Social      & 10,984  & 672,447   & - \& 172                & True      & 1 month       \\
MOOC        & Interaction & 7,144   & 411,749   & – \& 4                  & True      & 17 months     \\
LastFM      & Interaction & 1,980   & 1,293,103 & – \& –                  & True      & 1 month       \\
Enron       & Social      & 184     & 125,235   & – \& –                  & False     & 3 years       \\
Social Evo. & Proximity   & 74      & 2,099,519 & – \& 2                  & False     & 8 months      \\
UCI         & Social      & 1,899   & 59,835    & – \& –                  & False     & 196 days      \\
Flights     & Transport   & 13,169  & 1,927,145 & – \& 1                  & False     & 4 months      \\
Can. Parl.  & Politics    & 734     & 74,478    & – \& 1                  & False     & 14 years      \\
US Legis.   & Politics    & 225     & 60,396    & – \& 1                  & False     & 12 congresses \\
UN Trade    & Economics   & 255     & 507,497   & – \& 1                  & False     & 32 years      \\
UN Vote     & Politics    & 201     & 1,035,742 & – \& 1                  & False     & 72 years      \\
Contact     & Proximity   & 692     & 2,426,279 & – \& 1                  & False     & 1 month       \\ \bottomrule
\end{tabular}
}
\end{table*}

\subsection{Descriptions of Baselines} 
\label{sec:baseline}
We select the following ten baselines:\\

\textbf{JODIE}\cite{kumar2019predicting} focuses on bipartite networks of instantaneous user-item interactions. It employs two coupled RNNs to update the representation of the users and items recursively. A projection operation is introduced to learn the future representation trajectory of each user/item.

\textbf{DyRep}\cite{dyrepLearningRepresentationsDynamicGraphs} has a custom RNN that updates node representations upon observation of a new edge. For obtaining the neighbor weights at each time, DyRep uses a temporal attention mechanism, which is parameterized by the recurrent architecture.

\textbf{TGAT}\cite{xu2020inductive} computes the node representation by aggregating features from each node’s temporal-topological neighbors based on the self-attention mechanism. It is also equipped with a time encoding function for capturing temporal patterns.

\textbf{TGN}\cite{DBLP:journals/corr/abs-2006-10637} maintains an evolving memory for each node and updates this memory when the node is observed in an interaction, which is achieved by the message function, message aggregator, and memory updater. An embedding module is leveraged to generate the temporal representations of nodes.

\textbf{CAWN}\cite{wang2021inductive} first extracts multiple causal anonymous walks for each node, which can explore the causality of network dynamics and generate relative node identities. Then, it utilizes recurrent neural networks to encode each walk and aggregates these walks to obtain the final node representation.

\textbf{EdgeBank}\cite{poursafaeitowards} is a pure memory-based approach for transductive dynamic link prediction. It stores the observed interactions in the memory unit and updates the memory through various strategies. An interaction will be predicted as positive if it is retained in the memory and negative otherwise. 

\textbf{TCL}\cite{wang2021tcl} generates each node’s interaction sequence by performing a breadth-first search algorithm on the temporal dependency interaction sub-graph. Then, it presents a graph transformer that considers both graph topology and temporal information to learn node representations. It also incorporates a cross-attention operation for modeling the interdependencies of two interaction nodes.

\textbf{GraphMixer}\cite{cong2023we} shows that a fixed-time encoding function performs better than the trainable version. It incorporates the fixed function into a link encoder based on MLP-Mixer to learn from temporal links. A node encoder with neighbor mean-pooling is employed to summarize node features.

\textbf{NAT}\cite{luo2022neighborhood} adopts a novel dictionary-type neighborhood representation to gather the temporal neighbors of each node, and it then uses a recurrent process to learn the node representation from the historical neighbors of the current node and an RFF-based time embedding. The method constructs the query-induced subgraph without using neighbor samples to reduce computation costs.

\textbf{DyGFormer}\cite{yu2023towards} is a Transformer-based architecture for dynamic graph learning that only picks up information from previous first-hop interactions between nodes. It creates a neighbor co-occurrence encoding scheme based on a patching method and feeds them to the Transformer. This method enables the model to benefit from longer histories by exploring the correlations of the source node and destination node based on their sequences.

\begin{table*}[htbp]
\caption{Statistics of the TGB datasets}
\label{tab:tgb_statistics}
\resizebox{0.6\linewidth}{!}{
\begin{tabular}{@{}c|ccccc@{}}
\toprule
Datasets    & Domains     & \#Nodes & \#Links   & \#Steps& \#Surprise\\ 
\hline
tgbl-wiki   & Interaction      & 9,227   & 157,474   &    152,757  &    0.108    \\
tgbl-review      & Rating      & 352,637  & 4,873,540   & 6,865   &    0.987       \\
tgbl-coin        & Transaction & 638,486   & 22,809,486   & 1,295,720    &    0.120           \\
tgbl-comment      & Social & 994,790   & 44,314,507 & 30,998,030    &    0.823      \\
\bottomrule
\end{tabular}
}
\vspace{-0.2cm}
\end{table*}

\subsection{Performance on TGB} 
\label{sec:TGB}

Recently, researchers have noticed the size of benchmark datasets for the dynamic link prediction task is relatively small and proposed a large-scale temporal graph benchmark (TGB\footnote{\url{https://tgb.complexdatalab.com}}), we tested our method with the code provided by DyGLib-TGB\footnote{\url{https://github.com/yule-BUAA/DyGLib_TGB}}. Because the validation of the datasets takes days for each model, we directly use the baseline performance reported by DyGLib-TGB. Besides, due to the downloading issue in tgbl-flight, we do not compare our method on that dataset. The statistics of the datasets are listed in Table \ref{tab:tgb_statistics}

We have two key observations from Table \ref{tab:mrr_dynamic_link_property_prediction}. 
Firstly, CNE-N achieves the best performance on all four datasets against baseline models, even on datasets with low neighbor node re-interact probability (bad performance with EdgeBank, the model ), indicating co-neighbor encoding can effectively describe the evolution of the dynamic graph rather than the re-occurrence of the neighbor node. Secondly, we compare the efficiency between our method and two efficient baselines Jodie and GraphMixer in the largest two datasets tgbl-coin and tgbl-comment, Jodie takes 124 hours and 63 hours for one epoch of the training set, GraphMixer takes 2 hours, and 4 hours, and CNEN takes 0.67 hours and 1.42 hours, indicating the efficiency of our method on large scale dataset.

\begin{table}[htbp]
\centering
\caption{MRR for dynamic link property prediction, where Val is the abbreviation of Validation.}
\label{tab:mrr_dynamic_link_property_prediction}
\resizebox{\linewidth}{!}
{
% \setlength{\tabcolsep}{0.5mm}
% {
\begin{tabular}{c|c|cccc}
\hline
        Sets                     & Methods                 & tgbl-wiki  & tgbl-review  & tgbl-coin & tgbl-comment \\ \hline
\multirow{13}{*}{Val} & JODIE                   &  71.42 $\pm$ 0.76          &  \textbf{34.76 $\pm$ 0.06}    &   -        &       -        \\
                             & DyRep                   &  59.38 $\pm$ 1.82          &  \underline{33.85 $\pm$ 0.18}  &    {51.20 $\pm$ 1.40}         &    {29.10 $\pm$ 2.80}                \\
                             & TGAT                   &  65.14 $\pm$ 1.22    &    17.24 $\pm$ 0.89       &   60.47 $\pm$ 0.22        &     50.73 $\pm$ 2.47              \\
                             & TGN                    &  73.80 $\pm$ 0.39        &     33.17 $\pm$ 0.13    &  {60.70 $\pm$ 1.40}         &    {35.60 $\pm$ 1.90}                \\
                             & CAWN                   &  75.36 $\pm$ 0.34       &     {20.00 $\pm$ 0.10}      &     -      &         -     \\
                             & EdgeBank$_\infty$       &  56.13 $\pm$ 0.00  &  2.29 $\pm$ 0.00  &      31.54 $\pm$ 0.00       &    {10.87 $\pm$ 0.00}          \\
                             & EdgeBank$_\text{tw-ts}$  &  66.51 $\pm$ 0.00  &  2.90 $\pm$ 0.00  &     49.67 $\pm$ 0.00      &     {12.44 $\pm$ 0.00}               \\
                             & TCL                   &  {80.82 $\pm$ 0.14}   &   17.99 $\pm$ 1.72     &     66.85 ± 0.27        &      {65.10 $\pm$ 0.67}       \\
                             & GraphMixer          &  63.87 $\pm$ 0.53        &    28.28 $\pm$ 2.07       &    {70.38 $\pm$ 0.40}        &     \underline{70.19 $\pm$ 0.23}\\
                             & NAT             &  77.30 $\pm$ 1.10       &    30.20 $\pm$ 1.10	      &      -    &  -         \\
                             & DyGFormer             &  {\ul 81.62 $\pm$ 0.46}       &    21.92 $\pm$ 1.74      &    {\ul 72.97 $\pm$ 0.23}       &      61.33 $\pm$ 0.27        \\ 
                             & CNE-N             &  \textbf{82.29 $\pm$ 0.12}       &    19.70 $\pm$ 0.18      &    \textbf{74.39 $\pm$ 0.24}       &      \textbf{73.21 ± 0.21}        \\

                             \hline

\multirow{13}{*}{Test}       & JODIE                 &  63.05 $\pm$ 1.69        &   \textbf{41.43 $\pm$ 0.15}     &      -     &       -        \\
                             & DyRep                  &  51.91 $\pm$ 1.95     &     \underline{40.06 $\pm$ 0.59}      &     {45.20 $\pm$ 4.60}      &    {28.90 $\pm$ 3.30}             \\
                             & TGAT                   &    59.94 $\pm$ 1.63    &    19.64 $\pm$ 0.23       &   60.92 $\pm$ 0.57        &    56.20 $\pm$ 2.11                \\
                             & TGN                     &  68.93 $\pm$ 0.53       &    37.48 $\pm$ 0.23       &    {58.60 $\pm$ 3.70}        &   {37.90 $\pm$ 2.10}                   \\
                             & CAWN                     &  73.04 $\pm$ 0.60      &     {19.30 $\pm$ 0.10}      &    -       &       -         \\
                             & EdgeBank$_\infty$      &  52.50 $\pm$ 0.00  &  2.29 $\pm$ 0.00  &      35.90 $\pm$ 0.00       &   {12.85 $\pm$ 0.00}                   \\
                             & EdgeBank$_\text{tw-ts}$ &  63.25 $\pm$ 0.00   &  2.94 $\pm$ 0.00  &      57.36 $\pm$ 0.00      &   {14.94 $\pm$ 0.00}                 \\
                             & TCL                    &  {78.11 $\pm$ 0.20}  &  16.51 $\pm$ 1.85  &     68.66 $\pm$ 0.30          &    {70.11 $\pm$ 0.83}        \\
                             & GraphMixer              &  59.75 $\pm$ 0.39      &     36.89 $\pm$ 1.50     &   \underline{75.57 $\pm$ 0.27}         &      \underline{76.17 $\pm$ 0.17}      \\
                             & NAT             &  74.90 $\pm$ 1.00       &    34.10 $\pm$ 2.00	      &     -    &       -   \\
                             & DyGFormer               &  \underline{79.83 $\pm$ 0.42}     &     22.39 $\pm$ 1.52     &    {75.17 $\pm$ 0.38}       &      67.03 $\pm$ 0.14      \\ 
                             & CNE-N             &  \textbf{80.24 $\pm$ 0.20}       &    26.12 $\pm$ 0.25      &    \textbf{77.24 $\pm$ 0.21}       &     \textbf{78.97 ± 0.14}    
                             \\ 

                             \hline
\end{tabular}
% }
}
% \vspace{-0.4cm}
\end{table}

\end{document}